\documentclass[10pt,onecolumn]{IEEEtran}

\IEEEoverridecommandlockouts

\usepackage{fancyhdr}
\usepackage{cite}
\usepackage{amsmath,amssymb,amsthm,amsfonts}

\usepackage[linesnumbered,ruled]{algorithm2e} 
\usepackage{graphicx}
\usepackage{textcomp}
\usepackage{xcolor}
\def\BibTeX{{\rm B\kern-.05em{\sc i\kern-.025em b}\kern-.08em
    T\kern-.1667em\lower.7ex\hbox{E}\kern-.125emX}}
\usepackage{booktabs} 
\usepackage{hyperref}  
\usepackage{wrapfig}

\newtheorem{theorem}{Theorem}
\newtheorem{theorem*}{Theorem}
\newtheorem{lemma}{Lemma}
\newtheorem{proposition}{Proposition}
\newtheorem{corollary}{Corollary}

\newtheorem{example}{Example}

\begin{document}

\title{ 
Composite Neural Network: Theory and Application to PM2.5 Prediction }

\author{
\IEEEauthorblockN{
Ming-Chuan Yang
and Meng Chang Chen  }\\
\thanks{This work is accepted by IEEE Transactions on Knowledge and Data Engineering (TKDE).}

\IEEEauthorblockA{ \textit{Institute of Information Science, Academia Sinica}, Taiwan}\\
\{mingchuan,mcc\}@iis.sinica.edu.tw}

\maketitle

\begin{abstract}
This work investigates the framework and statistical performance guarantee of the composite neural network, which is composed of a collection of pre-trained and non-instantiated neural network models connected as a rooted directed acyclic graph, for solving complicated applications. A pre-trained neural network model is generally well trained, targeted to approximate a specific function.  The advantages of adopting a pre-trained model as a component in composing a complicated neural network are two-fold. One is benefiting from the intelligence and diligence of domain experts, and the other is saving effort in data acquisition as well as computing resources and time for model training.
Despite a general belief that a composite neural network may perform better than any a single component, the overall performance characteristics are not clear. In this work, we propose the framework of a composite network, and prove that a composite neural network performs better than any of its pre-trained components with a high probability. 

In the study, we explore a complicated application---PM2.5 prediction---to support the correctness of the proposed composite network theory. In the empirical evaluations of PM2.5 prediction, the constructed composite neural network models perform better than other machine learning models.
\end{abstract}


\section{Introduction}
Deep learning has seen great success in dealing with natural signals such as images and voices as well as 
  artificial signals such as natural language,   
whereas it is still in the early stages of handling complicated social and natural applications shaped by diverse factors (e.g., stock market prediction~\cite{DBLP:journals/tois/FengHWLLC19}) or that result from complicated natural processes (e.g., PM2.5 pollution level prediction~\cite{yi2020predicting}). Common to these complicated applications is their unbounded applicable data sources, 
which may not be available all at once, 
and their processes, which are difficult to learn from limited data. Consequently, their neural network based solutions often require frequent revisions as more relevant data are available or more data is made available, or the understanding of the process is enhanced. Although neural networks can approximate arbitrary functions~\cite{Hornik1991approximation}, competent neural networks for complicated applications are unrealistic for the above reasons, which motivates this study to devise an effective, realistic approach for such applications.

The obvious drawbacks of traditional approaches to suitable neural network models include a lack of flexibility given new data sources and knowledge, difficulty in improving problem modeling and decomposition, and an inability to employ the proven efforts of others. The main idea of the proposed composite neural network is to compose several neural network models, especially pre-trained models (i.e., neural network models with instantiated weights), based on the availability of data and domain knowledge, to solve complicated applications.  

An emerging trend in deep learning solution development is to employ well-crafted pre-trained neural networks, especially for use as a specific function/component to synthesize a neural network model. Many popular pre-trained neural network models are fine-tuned with adequate training data and made available to the public either as open-source or commercial products. In practice, training a large neural network is infeasible due to the limitations of computing resources. Pre-trained components may alleviate the problem by decomposing the problem into several sub-problems, each of which can be solved by a neural network component which can be trained separately. The advantages of adopting a pre-trained model in composing a complicated neural network are two-fold. One is benefiting from the intelligence and diligence of domain experts, and the other is saving effort in data acquisition as well as computing resources and time for model training. 

\begin{figure}
\centering{
\includegraphics[width=0.4\textwidth]{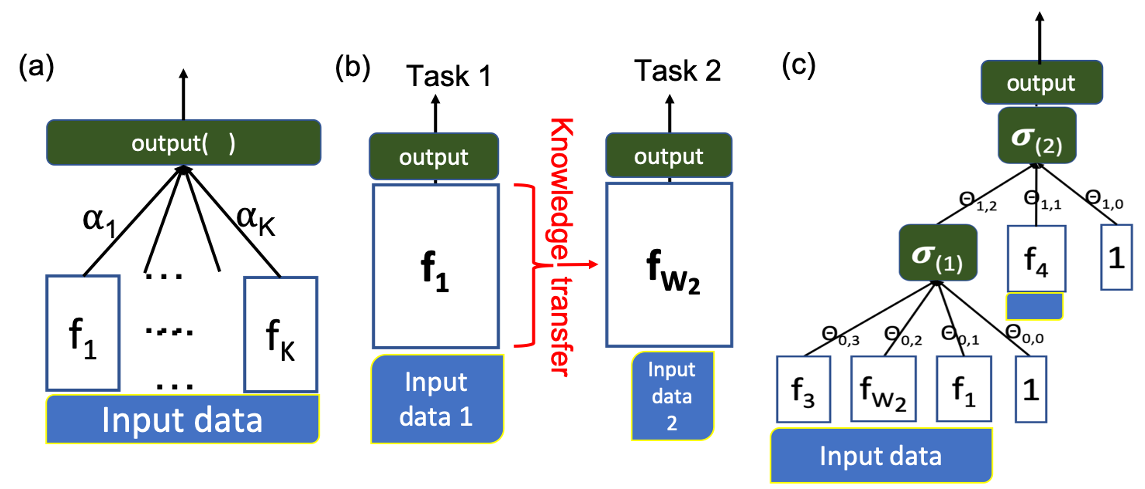}
}
\caption{Illustrations of (a) ensemble learning, (b) transfer learning, and (c) composite neural network.} 
\label{ExampleOfEnsembleTransfer}
\end{figure}

During the training phase of a composite network, the weights of pre-trained models are frozen to maintain their original quality, and to save training time for less trainable parameters, whereas the weights of their incoming and outgoing edges are trainable. Note that a user may choose the weights of a pre-trained component trainable for their particular purpose. For instance, in transfer learning, the weights of the pre-trained network may be used as initial values in the training phase of the overall neural network. 
Ensemble learning~\cite{freund1997decision} and transfer learning~\cite{galanti2016theoretical} both apply additional data and neural network models to improve accuracy. In deep learning, ensemble learning (Fig.~\ref{ExampleOfEnsembleTransfer}(a)) employs multiple neural networks together to make decisions whereas transfer learning (Fig.~\ref{ExampleOfEnsembleTransfer}(b)) applies knowledge learned from other neural networks to assist in solving the original problem.  
Although all the models consider pre-trained components, the ensemble learning basically adopts homogeneity learners (i.e., learners for the same function), and transfer learning methods are applied in the situation of insufficient training data. However, the proposed composite neural network is mainly for the incorporation of solutions of sub-problems, regardless heterogeneous not.    
\begin{figure}
\centering{
\includegraphics[width=0.43\textwidth]{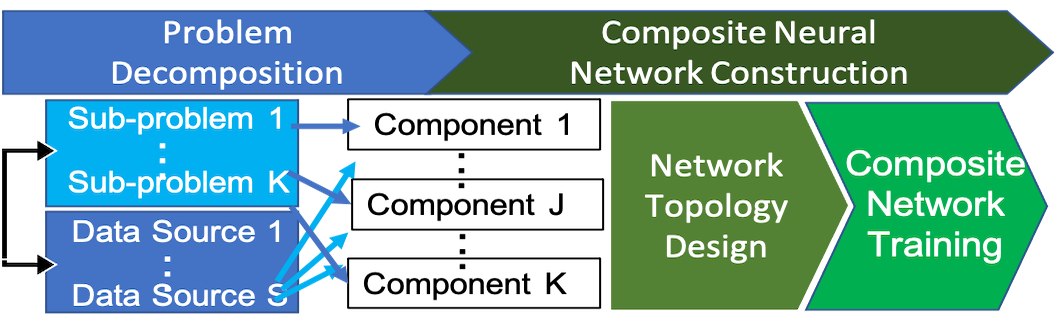}
}
\caption{Framework of Composite Neural Network Construction} 
\label{Framework of Decomposition-Composition}
\end{figure}

Ensemble learning and transfer learning have their constraints. Ensemble learning ensembles learners that must have the accuracy $>50\%$~\cite{zhou2012ensemble}. Transfer learning~\cite{galanti2016theoretical} assumes, for a source-target domain pair, there is an intermediate representation that can be transferred for domain adaptation. Unlike ensemble learning or transfer learning, the proposed composite network framework allows a generic condition with a statistical performance guarantee. In addition, in the literature, some negative effects have been observed, e.g., Zhou et al.~\cite{zhou2002ensembling} pointed out that “many could be better than all” in the typical ensemble settings, and  Chen et al.~\cite{chen2019catastrophic} showed the transfer learning has suffered from the “negative transfer” problem.
The papers of D{\v{z}}eroski et al.~\cite{dvzeroski2004combining} and of Gashler et al.~\cite{gashler2008decision} also concluded that an ensemble is not always strictly better than its best component because of the low diversity between members. 
These facts imply that the claim "the more components, the better performance" may be not always true.
The gap of theoretical analysis that supports or opposes the claim motivates this study.

PM2.5 (particulate matter with a diameter less than 2.5 $\mu m$)
has become a great concern due to its proven threat to human health~\cite{turner2017ambient}. 
PM2.5 is a collection of aerosol material primarily composed of ammonium sulfate, ammonium nitrate, organic carbonaceous mass, elemental carbon, and crustal mineral material emitted from sources such as vehicles, power plants and factories, fossil fuel burning, construction, farming activities, sea salt and dust, and remote transportation{~\cite{wei2016transfer,yi2018deep,li2020integrating,yi2020predicting}}. Both the constituents and sources of PM2.5 vary from one location to the other{~\cite{wei2016transfer,yi2018deep}}, from
one season to the other{~\cite{li2020integrating,yi2020predicting}}. For instance, for the seaside rural areas, dust and sea salt are the major causes, while in industrialized countries, fossil fuel burning is the major source. 
Therefore, PM2.5 prediction must be temporally and spatially dependent.
The life cycle and dispersion of PM2.5 depend on factors such as the type of PM2.5, weather conditions, terrain context, and chemical transformations that furthermore complicate the PM2.5 prediction~\cite{yi2020predicting}. As a result, predicting the PM2.5 level in the next few hours for a particular area is a great challenge. 

In this paper, we answer the challenge of solving complicated applications, and propose a framework and construction algorithms for a composite neural network. 
Then we use the complicated application -- PM2.5 prediction to demonstrate the efficacy of the composite neural network and its applicability to complicated real-world problems. As illustrated in Fig \ref{Framework of Decomposition-Composition}, first, a complicated application is decomposed into subtasks, and then some of them are selected as the candidates of pre-trained components. Once the pre-trained components are trained separately or obtained elsewhere, and the topology is decided, then an end-to-end training is performed to construct the final composite neural network.

The contributions in this paper are the following.  (1) We propose a framework for the composite neural network, and provide a theoretical analysis of statistical performance guarantee. 
(2) We provide two heuristic algorithms with alternative composite neural network design principles for performance comparison. (3) We empirically evaluate the performance of composite neural network algorithms and several traditional machine learning methods on PM2.5 prediction data sets; the outcomes support the proposed theory.

The remainder of this paper is organized as follows. We introduce the composite neural network in Section 2, and analyze its performance bounds in Section 3. Section 4 includes several algorithms for composite neural network construction. Section 5 shows intensive evaluations of various composite neural network constructions and traditional machine learning methods, and their comparisons. We discuss related work in Section 6 and the issues discovered during this study in Section 7.

\section{Concept of Composite Neural Network}

A typical single-layer neural network can be presented as
$f_{\sigma,\mathbf{W_1}}(\mathbf{x})$=$w_{1,1}\sigma\left(\sum_{i=1}^{d} w_{0,i} \mathbf{x}_i+w_{0,0} \right)+w_{1,0}$,
where $\mathbf{x}$ is the input vector, $\mathbf{W}_1$ is the matrix of weights, and $\sigma$ is the activation function. In this work we consider differentiable activation functions $\sigma: \mathbb{R}\to \mathbb{R}$ such as 
the the logistic function $\sigma(z)=1/(1+e^{-z})$ 
and the hyperbolic tangent $\sigma(z)=(e^{z}-e^{-z})/(e^{z}+e^{-z})$.    
If there is no ambiguity on the activation function, the $\sigma$ function is skipped to simplify notation and the neural network is denoted as $f_{\mathbf{W}}(\mathbf{x})$.

A composite neural network (also termed a composite network) is composed of a set of pre-trained and non-instantiated neural network models that form a directed acyclic graph. For a pre-trained model, its weight matrix ${\mathbf{W}_j}$ is fixed after its original training process, denoted as $f_{j}$ to distinguish it from a non-instantiated network. A non-instantiated network is denoted as $f_{\mathbf{W}_j}$; its weights ${\mathbf{W}_j}$ are not determined until the completion of the training process of the whole composite neural network. Both pre-trained and non-instantiated networks are called components of a composite neural network. 

\begin{table}[h]
    \centering
    \caption{Summary of Notations}\label{tab:notations}
    \begin{tabular}{r|l}
    \toprule
      notation   &  definition\\
    \midrule
      $\mathbf{W}$  & a matrix of weights in a neural network  \\
      $\sigma(z)$   & activation functions in a neural network \\
      $[N]$ & $\{1,...,N\}$; further, $[K]^{+}\triangleq\{0,1,...,K\}$  \\
    $\{(\mathbf{x}^{(i)},y^{(i)})\}_{i\in[N]}$   & a set of $N$ input-label pairs \\
    $f_{\sigma,\mathbf{W}}(\mathbf{x})$   &  a neural network defined by $\sigma$ and $\mathbf{W}$\\
    $\{f_{j}(\mathbf{x})\}_{j\in[K_1]}$     &  a set of $K_1$ pre-trained networks; $K_1\geq 1$  \\
    $\{f_{\mathbf{W}_j}(\mathbf{x})\}_{j\in[K_2]}$      & a set of $K_2$ non-instantiated networks \\
      $\{h_j(\mathbf{x})\}_{j\in[K]}$    &  $\{f_{j}(\mathbf{x})\}_{j\in[K_1]}\cup\{f_{\mathbf{W}_j}(\mathbf{x})\}_{j\in[K_2]}$\\
       $\mathbf{\Theta}$  & a matrix of weights in a composite network  \\
     $g_{\mathbf{\Theta}}(h_1,...,h_K)$  & an $r$-layer composite network of $h_j$s by $\sigma$, $\mathbf{\Theta}$:\\
     &\scriptsize{ $L_{\Theta_{(r+1)}}\left(\sigma_{(r+1)}\left( \cdots \sigma_{(1)}\left(L_{\Theta_{(0)}} \left(h_1,...,h_K\right)\right)\right)\right)$ }\\
     $L(\Theta; h_1,...,h_K)$  & $\sum_{j=0}^{K}{ \theta_jh_j(\mathbf{x})}$; $h_0=1$, linear combination \\
        $\vec{h}_j$  &  $(h_j(\mathbf{x}^{(1)}),\cdots,h_j(\mathbf{x}^{(N)}))$;  $\vec{h}_0\triangleq \vec{1}$  \\
       $\vec{e}_j$  & an unit vector in the standard basis of $\mathbb{R}^{K+1}$ \\
        $\mathcal{B}_{K+1}$ &  $\{\vec{e}_j\}_{j\in[K]^{+}}$ \\
    \bottomrule
    \end{tabular}
\end{table}

For a given set of $K$ components $\{h_j(\mathbf{x})\}_{j=1}^{K}$, each component $h_j$, which can be pre-trained or non-instantiated, has an input vector $\mathbf{x}$ and an output vector $\mathbf{y}_j$. Let  $h_0$ be the constant function $1$. Then the linear combination with a bias $\Theta=(\theta_0,\theta_1,\dots,\theta_K)$ is defined as $L(\Theta; h_1,...,h_K)=\sum_{j=0}^{K}{ \theta_jh_j(\mathbf{x})}$. When $\Theta$ is learned in the training phase, the composite network is denoted as $L_{\Theta}(h_1,...,h_K)$.
To extend the notation further, a neural network with $h$ hidden layers is denoted as {\small $L_{\Theta_{(h+1)}}\left(\sigma_{(h+1)}\left( \cdots \sigma_{(1)}\left(L_{\Theta_{(0)}} \left(h_1,...,h_K\right)\right)\right)\right)$}\normalsize, illustrated as in Fig.~\ref{Framework of Decomposition-Composition}(c),  where the braced number in the subscript indicates the layer number. The components can be in any layer and its output can be fed to any components in the upper layers. Example \ref{example of CNN} shows an example composite network.

\begin{example}
\label{example of CNN}
A composite neural network
 $\sigma_{(2)} ( \theta_{1,0}+ \theta_{1,1} f_{4}(\mathbf{x}_4)+  \theta_{1,2}\sigma_{(1)}( \theta_{0,0}+ \theta_{0,1} f_{1}(\mathbf{x}_1)+ $ $\theta_{0,2} f_{\mathbf{W}_2}(\mathbf{x}_2)+  \theta_{0,3} f_{3}(\mathbf{x}_3)) )$, as depicted in Fig.~\ref{Framework of Decomposition-Composition}(c), 
 can be denoted as 
 $\sigma_{(2)}\left( L_{(1)}\left(f_{4}, \sigma_{(1)} \left( L_{(0)}(f_{1},f_{\mathbf{W}_2},f_{3}) \right)\right)\right)$, with  $\mathbf\Theta$s removed for simplicity.
\end{example}

We assume that the training algorithm of the composite network is the stochastic gradient descent backpropagation algorithm and the loss function is the $L_2$-norm of the difference vector. 
The loss function for a trained composite neural network $g_{\mathbf{\Theta}}$ is defined as 
\begin{equation}
\mathcal{E}_{\mathbf{\Theta}}\left(\mathbf{x};g_{\mathbf{\Theta}}\right)=\frac {\langle g_{\mathbf{\Theta}}\left(\mathbf{x}\right)-\vec{y}, g_{\mathbf{\Theta}}\left(\mathbf{x}\right)-\vec{y} \rangle}{N},
\end{equation}
where $\langle {\cdot} ,{\cdot} \rangle$ is the standard inner product and $\vec{y}$ is the ground truth.
$\mathcal{E}_{\mathbf{\Theta}}\left(\mathbf{x};g_{\mathbf{\Theta}}\right)$ may be shortened to $\mathcal{E}\left(g_{\mathbf{\Theta}}\right)$. Clearly, the total loss depends on the training data $\mathbf{x}$, the components defined by $\{h_j\}_{j=1}^{K}$, the output activation $\sigma$, and the weight vector $\mathbf{W}$. 
Define $\mathcal{E}(\mathbf{x};f_j)$ (shortened to $\mathcal{E}(f_j)$, if there is no ambiguity) 
as the loss function of a single component $f_j$. It is expected that a good composite network design has low L2 loss, in particular lower than all its pre-trained components. Therefore, the goal is to find a feasible $\mathbf{\Theta}$ such that it meets the ``No-Worse'' property, i.e., $\mathcal{E}\left(g_{\mathbf{\Theta}}\right)<\min_{j\in[K]}{\mathcal{E}(f_j)}$.

In the following section we will prove that in some reasonable conditions, with high probability, a composite network has strictly lower training L2 loss than all of its pre-trained components. The expectation of L2 loss of a composite network is also with high probability lower than the expectation of the loss of all its pre-trained components. Furthermore, we will show a multi-layer composite network of mixed non-instantiated and pre-trained models that also, with high probability, performs better than any of its pre-trained models.

\section{Theoretical Analysis}
In this section, we first analyze the loss functions of a single-layer composite network, and subsequently extend the analysis to a complicated composite network to explore the characteristics of the composite network. Due to limited space, only ideas and sketches of proof are presented in this section. For the complete  proof, please refer to the appendix in the supplementary material.

A composite network constructed from a given set of pre-trained components $\{f_j\}_{j=1}^{K}$ forms an acyclic directed graph, which can be represented by postorder tree traversal. 
Without loss of generality, we assume the dimension of the output vector of all components is 1 in the following proofs. 
We denote $[K]^{+}$ the set from 0 to $K$, $\vec{f}_0=\vec{1}$, and $\vec{f}_j=(f_j(\mathbf{x}^{(1)}),\cdots,f_j(\mathbf{x}^{(N)}))$ as the sequence of the status of $f_j$  with input data $\mathbf{x}$ during the training phase. Similarly, the representation of the ground truth is $\vec{y}:=(y^{(1)},\cdots,y^{(N)})$. Let $\vec{e}_j$ be an unit vector in the standard basis of $\mathbb{R}^K$ for $j\in[K]$, e.g., $\vec{e}_1=(1,0,\cdots,0)$ and
$\mathcal{B}_{K}:=\{\vec{e}_j\}_{j=1}^{K}$. 
By $C^1$-mapping (function) we mean the mapping is differentiable and its derivative is a continuous function. 

The following assumptions are the default conditions in the following proofs.
\begin{enumerate}
    \item[A1.] Linearly independent components assumption: \\ 
$\forall i\in[K]^{+}, \nexists \{\beta_j\}\subset \mathbb{R}, \mbox{ s.t. } \vec{f}_i=\sum_{j\in[K]\setminus \{i\}}{\beta_j\vec{f}_j}$. 
    \item[A2.] No perfect component assumption:\\ 
$\min_{j\in [K]}\left\{ \sum_{i\in[N]}{|f_j(\mathbf{x}^{(i)})-\mathbf{y}^{(i)} |} \right\}>0$. 
    \item[A3.] The activation function and its derivative are $C^1$-mappings (i.e., it is differentiable and its differential is continuous) and the derivative is non-zero at some points in the domain.
    \item[A4.] The number of components, $K$, is less than $2\sqrt{N}-1$, where $N$ is the size of the training data set.  
\end{enumerate}

\subsection{Single-Layer Composite Network} \label{Theory_training_rmse} 

The first theorem states that if a single-layer composite network satisfies the above five assumptions, it meets the ``No-Worse'' property with high probability.

\begin{theorem}\label{theorem1}
Consider a single-layer composite network $g(\mathbf{x})=L_{(1)}(\sigma(L_{(0)}(f_1,...,f_K)))(\mathbf{x})$.
Then with probability of at least $1-\frac{K+1}{\sqrt{N}}$ there exists $\mathbf{\Theta}=\{\Theta_1,\Theta_0\}$ s.t. $\mathcal{E}_{\mathbf{\Theta}}\left(\mathbf{x};g\right)<\min_{j\in[K]}{\mathcal{E}(f_j(\mathbf{x}))}$.
\end{theorem}

We discuss two cases of the activation $\sigma$.
\begin{itemize}
    \item Case 1: $\sigma$ is a linear function.
    \item Case 2: $\sigma$ is not a linear function.
\end{itemize}

(\textbf{Case 1}) 
$\sigma$ is a linear activation such that a single-layer composite network such as $L_{(1)}(\sigma(L_{(0)}(f_1,...,f_K)))$ can be rewritten as a linear combination with bias, i.e., $g_{\mathbb{\theta}}(\mathbf{x})=\sum_{j\in[K]^{+}}{ \theta_jf_j(\mathbf{x})}$ with a mean squared error of $\mathcal{E}_{\mathbf{\Theta}}\left(\mathbf{x};g\right)= \frac{1}{N}\sum_{i=1}^{N}(g_{\mathbf{\Theta}}(\mathbf{x}^{(i)})-\mathbf{y}^{(i)})^2$. 
Clearly, the composite network $g_{\mathbb{\theta}}$ should have a mean squared error equal to or better than any of its components $f_j$, as $g_{\mathbb{\theta}}$ can always act as its best component.
To obtain the minimizer $\mathbf{\Theta}^{*}$ 
 for the error $\mathcal{E}_{\mathbf{\Theta}}\left(\mathbf{x};g\right)$, we must compute the partial differential ${\partial \mathcal{E}_{\mathbf{\Theta}}}/{\partial{\theta}_j}$
for all ${j\in[K]^{+}}$. After some calculations~\cite{horn1990matrix}, we have Eq~(\ref{eq2}).
\begin{equation} \label{eq2}
\mathbf{\Theta}^{*} = \left[ \mathbf{\theta}_j\right]_{j\in[K]^{+}} 
= 
\left[ \langle \vec{f_i},\vec{f_j} \rangle \right]_{i,j\in[K]^{+}}^{-1}\times
\left[ \langle \vec{f_i},\vec{y} \rangle \right]_{i\in[K]^{+}}
\end{equation}
Since Assumption A1 holds, the inverse matrix $\left[ \langle \vec{f_i},\vec{f_j} \rangle \right]_{i,j\in[K]^{+}}^{-1}$ exists and can be written down concretely to obtain $\Theta^{*}$ as in Eq.~(\ref{eq2}). Lemma~\ref{lemma_1} summarizes the above arguments.

\begin{lemma}\label{lemma_1}
Set ${\Theta}^{*}$ as in Eq.~(\ref{eq2}); then
\begin{equation}\label{eqlemma1}
\mathcal{E}(g_{\mathbf{\Theta}^{*}})\leq \min_{j\in[K]^{+}}\{\mathcal{E}(f_j)\}.
\end{equation}
\end{lemma}

There is a $\leq$ constraint on the loss function $\mathcal{E}(g_{\mathbf{\Theta}^{*}})$ in Eq.~(\ref{eqlemma1}) that is replaced by $<$ and a probability bound. If $\mathbf{\Theta}^{*}$ is not a unit vector, it is obvious that  $\mathcal{E}(g_{\mathbf{\Theta}^{*}})$ must be less than any $\mathcal{E}(f_j)$. Therefore, we proceed to estimate the probability of $\mathbf{\Theta}^{*}=\vec{e}_{j^{*}}$, where $j^{*}\in[K]^{+}$.
\begin{equation}\label{eq4}
\forall i\in[K]^{+},
\frac{\partial \mathcal{E}}{\partial{\theta}_i}\\
\big\vert_{
\mathbf{\Theta}=\vec{e}_{j^{*}}}
= 2
\langle\vec{f}_{j^{*}}-\vec{y},\vec{f}_i \rangle \\
\end{equation}
Eq.~(\ref{eq4}) shows the gradient of the error function with respect to $\theta_i$ conditioned on $\mathbf{\Theta}^{*}=\vec{e}_{j^{*}}$, which is the inner products of the difference between ${f}_{j^{*}}$ (the output of $g_{\mathbf{\Theta}^{*}}$) and the ground truth $\vec{y}$, and the output of each pre-trained component $\vec{f}_i$.  
%
When the minimizer $\mathbf{\Theta}^{*}=\vec{e}_{j^{*}}$, all the differentials $\frac{\partial \mathcal{E}}{\partial{\theta}_i}$ must equal zero, i.e., $\langle\vec{f_{j^{*}}}-\vec{y},\vec{f_{i}} \rangle=0$, or $\vec{f_{j^{*}}}-\vec{y}$ is perpendicular to $\vec{f_{i}}$. 
The following Lemma~\ref{lemma_JL} is an implication from the proof of the Johnson-Lindenstrauss Lemma~\cite{johnson1984extensions}, which states that a randomly sampled unit vector $\vec{v}$ (denoted as $\Pr_{\vec{v}\in\mathbb{R}^{N}}$) is approximately perpendicular to a given vector $\vec{u}$ with high probability in a high dimensional space..

\begin{lemma} \label{lemma_JL}
For a large enough $N$ and given $\vec{u}\in\mathbb{R}^{N}$, there is a constant $c>0$, s.t. for $\eta=cos^{-1}(1-c/\sqrt{N})$,
\begin{equation}\label{JL}
\Pr_{\vec{v}\in\mathbb{R}^{N}}\left\{ | \angle_{\vec{u},\vec{v}} -\frac{\pi}{2}|\leq \eta\right\} \geq 1-\frac{1}{\sqrt{N}}
\end{equation}
where $\angle_{\vec{u},\vec{v}}$ is the angle between $\vec{u}$ and $\vec{v}$.
\end{lemma}
The complement of Eq.~(\ref{JL}) is
\begin{equation}\label{JL_complement}
\Pr_{\vec{v}\in\mathbb{R}^{N}}\left\{ | \angle_{\vec{u},\vec{v}} -\frac{\pi}{2}|{>} \eta\right\} {<} \frac{1}{\sqrt{N}}
\end{equation}
Note that angles $\angle_{\vec{y} ,\vec{f}}$ , $\angle_{\vec{f}-\vec{y},\vec{f}}$, and $\angle_{\vec{f}-\vec{y},-\vec{y}}$ are the three inner angles of the triangle such that 
$\angle_{\vec{y},\vec{f}}+\angle_{\vec{f}-\vec{y},\vec{f}}+\angle_{\vec{f}-\vec{y},-\vec{y}}=\pi $. 
From Lemma~\ref{lemma_JL}, as $\angle_{\vec{y},\vec{f}}$ is likely a vertical angle (i.e., $\pi/2$),  $\angle_{\vec{f}-\vec{y},\vec{f}}$ must be less likely to be a vertical angle, which implies  
$\Pr\{ \langle\vec{f}-\vec{y}, \vec{f} \rangle=0  \}\leq\Pr\{|\angle_{\vec{f}-\vec{y},\vec{f}}-\pi/2|<\eta \}$; thus, $\leq \Pr\{| \angle_{\vec{y},\vec{f}} -{\pi}/{2}|>\eta \}.$
The following Lemma~\ref{lemma_3} immediately follows Lemma~\ref{lemma_JL} and Eq.~(\ref{JL_complement}).
\begin{lemma}\label{lemma_3} 
Following Lemma~\ref{lemma_JL}, then for given $\vec{y}\in\mathbb{R}^N $, 
$$
\Pr_{\vec{f}\in\mathbb{R}^{N}}
\left\{ \langle\vec{f}-\vec{y},\vec{f} \rangle=0  \right\}< \frac{1}{\sqrt{N}}.
$$
\end{lemma}

Lemma~\ref{lemma_3} shows that the probability of the output of one component is perpendicular to the difference between itself and the ground truth. For $K$ components and a bias, Lemma~\ref{lemma_4} gives a worst bound.

\begin{lemma}\label{lemma_4}
$\Pr\left\{\mathcal{E}(g_{\mathbf{\Theta}^{*}})= \min_{j\in[K]^{+}}\{\mathcal{E}(f_j)\} \right\}< \frac{K+1}{\sqrt{N}}$, i.e., $\Pr\left\{\exists\Theta^*:\mathcal{E}(g_{\mathbf{\Theta}^{*}})< \min_{j\in[K]^{+}}\{\mathcal{E}(f_j)\} \right\}\geq 1-\frac{K+1}{\sqrt{N}}$.
\end{lemma}

(\textbf{Case 2}) $\sigma$ is not a linear function. The idea of the proof is to find an interval in the domain of $\sigma$ such that the output of $L_{(1)}(\sigma(\cdot))$ approximates a linear function as close as possible. 
This means there is a setting such that the non-linear activation function performs almost as well as the linear one; since the activation  $L_{(1)}(\sigma(\cdot))$ acts like a linear function, the lemmas of Case 1 are applicable. The conclusion of this case is stated as Lemma~\ref{lemma_case2}, while we introduce important properties in Lemmas~\ref{Inverse Function} and \ref{TaylorLagrange} for key steps in the proof. 

Since  $\sigma$ satisfies Assumption~A3,  the inverse function theorem 
of Lemma~\ref{Inverse Function} is applicable.

\begin{lemma} \label{Inverse Function} (Inverse function theorem~\cite{rudin1964principles})\label{IFThm}\\
Suppose $\mu$ is a $C^1$-mapping of an open set $E\subset \mathbb{R}^n$ to $\mathbb{R}^n$, $\mu'(z_0)$ in invertible for some $z_0\in E$, and $y_0=\mu(z_0)$.  (I.e., $\mu$ satisfies Assumption~A3.) Then \\
(1) there exist open sets $U$ and $V$ in $\mathbb{R}^n$ such that $z_0\in U$, $y_0\in V$, $\mu$ is one-to-one on $U$, and $\mu(U)=V$;\\
(2) if $\nu$ is the inverse of $\mu$, defined in $V$ by $\nu(\mu(x))=x$ for $x\in U$, then $\nu\in C^1(V)$.
\end{lemma}

We also need the following lemma as an important tool.

\begin{lemma} \label{TaylorLagrange}
(Taylor's theorem with Lagrange remainder~\cite{courant2012introduction})\\
If a function $\tau(y)$ has continuous derivatives up to the ($l+1$)-th order on a closed interval containing the two points $y_0$ and $y$, then
$$
\tau(y)=\tau(y_0)+\tau^{(1)}(y_0)(y-y_0)+\cdots+\frac{\tau^{(l)}(y_0)}{l!}(y-y_0)^l+R_l
$$
with the remainder $R_l$ given by the expression for some $c\in [0,1]$: 
$$
R_l=\frac{\tau^{(l+1)}(c(y-y_0))}{(l+1)!}(y-y_0)^{l+1}.
$$
\end{lemma}

Let $l=1$, $\tau(y)$ be obtained such that \footnotesize{
\begin{equation}\label{eq_8}
  \tau(y)=\tau(y_0)+\tau^{(1)}(y_0)(y-y_0)+\frac{\tau^{(2)}(c(y-y_0))} {2!}(y-y_0)^2.  
\end{equation}}\normalsize  
The second-degree term can be used to bound the approximation error.

Now we are ready to give more details to sketch the proof of Case 2. 
Denote $\mathbf{\Theta}_0^{*}$ as the minimizer of Case 1, i.e., the corresponding $g_{\mathbf{\Theta}_0^*}=L^*_{(0)}(f_1,...,f_K)$ satisfies $\mathcal{E}(g_{\mathbf{\Theta}_0^{*}})< \min_{j\in[K]^{+}}\{\mathcal{E}(f_j)\}=\mathcal{E}(f_{j^*})$ with high probability, and denote $\mathbf{\Theta}_{\epsilon}=\{\Theta_{1,\epsilon},\Theta_{0,\epsilon}\}$ corresponding to 
\begin{equation}\label{g theta}
g_{\mathbf{\Theta}_{\epsilon}}=L_{(1),\epsilon}(\sigma(L_{(0),\epsilon}(f_1,...,f_K))),
\end{equation}
called the scaled $\sigma$ function. 
Lemma~\ref{lemma_case2} below states a clear condition of a linear approximation of a non-linear activation function.

\begin{lemma}\label{lemma_case2}
For the given $g_{\mathbf{\Theta}^*_{0}}$, $\{\mathbf{x}^{(i)}\}_{i\in [N]}$, and any $0<\epsilon\leq 1$, there exists $\mathbf{\Theta}_{\epsilon}=\{\Theta_{1,\epsilon},\Theta_{0,\epsilon}\}$ such that
\begin{equation}\label{eq_Lemma7}
   \forall i\in[N], |g_{\mathbf{\Theta}_{\epsilon}}(\mathbf{x}^{(i)})-g_{\mathbf{\Theta}^*_{0}}(\mathbf{x}^{(i)})|<\epsilon. 
\end{equation}
Furthermore, for small enough $\epsilon$, 
\begin{equation}\label{eq_Lemma7_2}
 \Pr\left\{\mathcal{E}(g_{\mathbf{\Theta}_{\epsilon}})< \min_{j\in[K]^{+}}\{\mathcal{E}(f_j)\} \right\} \geq 1-\frac{K+1}{\sqrt{N}}.
\end{equation}
\end{lemma}

From the definition of $g_{\mathbf{\Theta}_{\epsilon}}$,
finding a proper $L_{(0),\epsilon}(\cdot)$ and $L_{(1),\epsilon}(\cdot)$ are the major steps in the proof of Eq.~(\ref{eq_Lemma7}). 
$L_{(0),\epsilon}(\cdot)$ maps the output range of $g_{\mathbf{\Theta}^*_0}(\mathbf{x})$ to an interval $(-\gamma+z_0,\gamma+z_0)\subset U_0$ for some $\gamma>0$  satisfying $\sigma'(z_0)\neq 0$. 
The scaling factors $M_0$ and $L_{(0),\epsilon}(\cdot)$ are defined as 
\begin{equation}\label{M0}
M_0 =\frac{2}{\gamma}{\max_{i\in[N]}\{|g_{\mathbf{\Theta}^*_0}(\mathbf{x}^{(i)})|\}}
\end{equation}
\begin{equation}
L_{(0),\epsilon}(\mathbf{x}) = M_0^{-1}g_{\mathbf{\Theta}^*_0}(\mathbf{x})+z_0. 
\end{equation}
It is clear that the range of $L_{(0),\epsilon}(\mathbf{x})$ falls within $U_0$. $L_{(1),\epsilon}(y)$ intends to map the output range of $\sigma$ back to $g_{\mathbf{\Theta}^*_0}(\cdot)$, and is defined as the expansion of $\tau(\cdot)$ following Eq.~(\ref{eq_8}) without the error term.
\begin{equation} \label{L1}
L_{(1),\epsilon}(y)= M_0\cdot\tau^{(1)}(y_0)\cdot y+
M_0\cdot\left(z_0-\tau^{(1)}(y_0)\cdot y_0 \right).
\end{equation}
Reversing the scaling and translating, Eq.~(\ref{L1}) can be rewritten as
\begin{equation}\label{g*}
M_0\left(\tau(y_0)+\tau^{(1)}(y_0)(\sigma\left( M_0^{-1}g_{\mathbf{\Theta}^*_0}(\mathbf{x})+z_0 \right)-y_0) \right)-z_0,
\end{equation}
which equals 
$g_{\mathbf{\Theta}^*_0}(\mathbf{x})$ plus an error bounded by   $M_0M_1\gamma^2$, where 
\footnotesize{
\begin{equation}\label{M1}
M_1=5\sup_{z\in U_0}\left\{ |\tau^{(2)}(\sigma(z)-\sigma(z_0))| \cdot\left(\frac{\sigma(z)-\sigma(z_0)}{z-z_0}\right)^2\right\}.
\end{equation} }\normalsize

The precise setting of $\gamma$ can be obtained from $M_0M_1\gamma^2<\epsilon$. 
Then, with $\gamma$ and the properties of Lemmas~\ref{Inverse Function} and \ref{TaylorLagrange}, it can be verified that $g_{\mathbf{\Theta}_{\epsilon}}(\mathbf{x}^{(i)})=L_{(1),\epsilon}(\sigma(L_{(0),\epsilon}(\mathbf{x}^{(i)})))$ fits Eq.~(\ref{eq_Lemma7}).

Eq.~(\ref{eq_Lemma7}) implies $(g_{\mathbf{\Theta}_{\epsilon}}(\mathbf{x}^{(i)})-y^{(i)})^2<(|g_{\mathbf{\Theta}^*_{0}}(\mathbf{x}^{(i)})-y^{(i)}|+\epsilon)^2$, which
can derive $\mathcal{E}(g_{\mathbf{\Theta}_{\epsilon}})< \mathcal{E}(g_{\mathbf{\Theta}_0^{*}})+\Delta(\epsilon)$, where $\Delta(\epsilon)$ is an increasing function of $\epsilon$ when the other parameters are fixed. Hence, if $\epsilon$ is small enough, we have  $\Delta(\epsilon)\leq \frac{\mathcal{E}(f_{j^*})-\mathcal{E}(g_{\mathbf{\Theta}^{*}_0})}{3}$. By further considering $\mathcal{E}(g_{\mathbf{\Theta}{\epsilon}})< \mathcal{E}(g_{\mathbf{\Theta}_0^{*}})+\Delta(\epsilon)$, it is easy to see that $\mathcal{E}(g_{\mathbf{\Theta}{\epsilon}})< \mathcal{E}(f_{j^*})$.  The probability of $\mathcal{E}(g_{\mathbf{\Theta}^*_0})<\mathcal{E}(f_{j^*})$ of Eq.~(\ref{eq_Lemma7_2}) can be inferred from Lemma~\ref{lemma_4} of Case 1.
Example~\ref{scaled fucntion} below shows how to construct a scaled activation function that satisfies Eq.~(\ref{eq_Lemma7}).

\begin{example} \label{scaled fucntion}
Here we take a logistic function $\sigma(z)=\frac{1}{1+e^{-z}}$ in the context of PM2.5 prediction to construct a scaled logistic function. 
Let notations $g_{\mathbf{\Theta}^*_0}(\cdot)$, $z_0$,
$U_0$, $V_0$, and 
$\tau(\cdot)$ be as previously defined.
The assumption that the highest PM2.5 measurement is less than 1000 (i.e., $\max_{i\in[N]}\{|g_{\mathbf{\Theta}^*_0}(\mathbf{x}^{(i)})|\}<1000$) fits the reality for most countries. 
Observe that $\sigma^{(1)}(0)=\frac{1}{4}$, $\sigma(0)=\frac{1}{2}$, and hence it is valid to set $z_0=0$. 
Consider $(-\gamma,\gamma)\subset [-1,1]$ and hence , $y_0=\sigma(0)$ and $y=\sigma(z)\in (0.25,0.75)$.

The inverse function of $\sigma(z)$ is $\tau(y)=\ln{\frac{y}{1-y}}$ for $y\in (0,1)$, which also can be represented as $\tau(y)=4y-2+\frac{\tau^{(2)}(c(y-y_0))}{2}(y-y_0)^2$ for some $c\in (0,1)$ by Lemma~\ref{TaylorLagrange}. 
From Eq. (\ref{M0}), the scaling factors  $M_0=2\gamma^{-1}{\max_{i\in[N]}\{|g_{\mathbf{\Theta}^*_0}(\mathbf{x}^{(i)})|\}} < 2\cdot 10^3 \gamma^{-1}$, and from Eq. (\ref{M1}), $M_1=$ 
$5\sup_{z\in U_0} \left\{\tau^{(2)}(\sigma(z)-\sigma(z_0))\left[\left(\sigma(z)-\sigma(z_0)\right)/(z-z_0)\right]^2 \right\}$, which is less than 50 for $z\in (-\gamma,\gamma)$.
From Eq. (\ref{g*}), the scaled logistic function as
$g_{\mathbf{\Theta}_{\epsilon}}(\mathbf{x})=M_0\cdot\left(4\sigma\left( M_0^{-1}g_{\mathbf{\Theta}^*_0}(\mathbf{x}) \right)-2 \right)$. 

Now we claim that  for any given $\epsilon\in (0,1]$, $g_{\mathbf{\Theta}^*_{0}}(\cdot)$ and $\{\mathbf{x}^{(i)}\}_{i\in [N]} $, we have $|g_{\mathbf{\Theta}_{\epsilon}}(\mathbf{x}^{(i)})-g_{\mathbf{\Theta}^*_{0}}(\mathbf{x}^{(i)})|<\epsilon$. Here is a short verification.  Observe $\forall i\in[N], M_0^{-1} g_{\mathbf{\Theta}^*_0}(\mathbf{x}^{(i)})\in (-\gamma,\gamma)$. Also, if $z\in (-\gamma,\gamma)$, then  $|\frac{\tau^{(2)}(c(y-y_0))}{2}|(y-y_0)^2<M_1\gamma^2$. 
Recall that $\tau\circ\sigma(\cdot)$ is an identity function, $y=\sigma(M^{-1} g_{\mathbf{\Theta}^*_0}(\mathbf{x}))$, and $|\tau(y)-(4y-2)|$ 
$<M_1 \gamma^2$. 
That is, $|M_0^{-1} g_{\mathbf{\Theta}^*_0}(\mathbf{x})-\left[4\sigma(M_0^{-1} g_{\mathbf{\Theta}^*_0}(\mathbf{x}))-2\right]|<M_1 \gamma^2$. Multiply by $M_0$ on both sides and replace the bracket term with $g_{\mathbf{\Theta}_{\epsilon}}(\mathbf{x})$; we have 
$|g_{\mathbf{\Theta}^*_{0}}(\mathbf{x})-g_{\mathbf{\Theta}_{\epsilon}}(\mathbf{x})|
<M_0M_1 \gamma^2<10^5\gamma$. Hence, setting $\gamma=10^{-5}\epsilon$ verifies this claim.
\end{example}
From Lemma~\ref{lemma_case2}, we can conclude that there exists $\mathbf{\Theta}_{\epsilon}$ such that a non-linear single-layer composite network performs at least as well as the linear case with arbitrary small error. Thus, the proof of Case 2 is concluded.
The proofs of Cases 1 and 2 above complete the proof of Theorem~\ref{theorem1}.

\subsection{Complicated Composite Network}
In the previous section, we investigated the performance of a single-layer composite network comprising several pre-trained components connected by an activation function. Now we consider expanding the composite network in terms of width and depth. Formally, for a given pre-trained component $f_K$ and a trained composite network $g_{K-1}$ of $K-1$ components $(f_1,...,f_{K-1})$, we study the following two questions in this section.
\begin{itemize}
    \item[Q1:] (Adding width) By adding a new pre-trained component $f_K$, we define $g_{K}=L_{(1)}(\sigma(L_{(0)}(f_1,...,f_{K-1},f_K ))$. Is there $\Theta$ such that $\mathcal{E}(g_{K-1}) > \mathcal{E}_{\Theta}(g_{K})?$ 
    \item[Q2:] (Adding depth) By adding a new pre-trained component $f_K$, let $g_{K}=L_{(K)}(\sigma(L_{(K-1)}(g_{K-1},f_K))$. Is there $\Theta$ such that $\mathcal{E}(g_{K-1}) > \mathcal{E}_{\Theta}(g_{K})?$
\end{itemize}
Lemma~\ref{AddOne} answers Q1, and  we require Proposition \ref{AddOne_Kis2} as the base of induction to prove it.
\begin{lemma}\label{AddOne}
Set $g_{K}=L_{(1)}(\sigma(L_{(0)}((f_1,...,f_{K-1},f_K )))$. With probability of at least $1-\frac{K+1}{\sqrt{N}}$,
there is $\mathbf{\Theta}$ s.t. $\mathcal{E}\left(g_{K-1}\right)> \mathcal{E}_{\mathbf{\Theta}}\left(g_{K}\right)$.
\end{lemma}
\begin{proposition}\label{AddOne_Kis2}
Consider the case of only two pre-trained models $f_0$ and $f_1$. There exists $(\alpha_0,\alpha_1)\in \mathbb{R}^2$ s.t.
\small{
$$
 \sum_{i\in[N]}{(f_1(\mathbf{x}^{(i)})-y^{(i)})^2}>\sum_{i\in[N]}{\left(\alpha_0 f_0(\mathbf{x}^{(i)})+\alpha_1 f_1(\mathbf{x}^{(i)})-y^{(i)}\right)^2}
$$}\normalsize
with a probability of at least $1-\frac{2}{\sqrt{N}} $.
\end{proposition}\label{proposition}

Proposition~\ref{AddOne_Kis2} can be proved by solving the inequality directly for the case of $K=2$, and then generalizing the result to larger $K$ by induction with the help of Lemma~\ref{lemma_3} to prove Lemma~\ref{AddOne}. 
Adding a new component $f_K$ to a composite network $g_{K-1}$ as in Q2, the depth of resulting $g_K$ increments by 1. If $\vec{g}_{K-1}$ and $\vec{f}_K$ satisfy A1 and A2, consider $\{g_{K-1},f_K\}$ as a new set of $\{f_1,f_2\}$ in the same layer. Consequently, we can apply the arguments in Case 2 of Theorem \ref{theorem1} to show 
Lemma~\ref{addDeep} in the following, which answers Q2 and says the resulting $g_K$ has a minimizer $\mathbf{{\Theta}^*}$ such that with high probability the loss decreases.
\begin{lemma}\label{addDeep}
Set $g_{K}=L_{(1)}(\sigma(L_{(0)}((g_{K-1},f_K))$.  If $\vec{g}_{K-1}$ and $\vec{f}_K$ satisfy A1 and A2, then with a probability of at least $1-\frac{2}{\sqrt{N}}$, there is $\mathbf{\Theta}$ s.t.  $\mathcal{E}\left(g_{K-1}\right)> \mathcal{E}_{\mathbf{\Theta}}\left(g_{K}\right)$.
\end{lemma}

The proof of Lemma~\ref{addDeep} is similar to the proof of Case 2 in the previous sub-section.
Lemmas~\ref{AddOne} and \ref{addDeep} imply a greedy strategy to build a complicated composite network.
Recursively applying both lemmas, we can build a complicated composite network as desired. Theorem~\ref{DeeperWider} gives a formal statement of the constructed complicated composite network with a probability bound. The proof of Theorem~\ref{DeeperWider} is based on mathematical induction on layers and the worst case probability is over-estimated by assuming each layer could have up to $K$ components.

\begin{theorem}\label{DeeperWider}
For an $H$-hidden layer composite network with $K$ pre-trained components, there exists $\mathbf{\Theta}^*$ s.t.  
\small $$
\mathcal{E}_{\mathbf{\Theta}^*}(g)<\min_{j\in[K]^+}\{\mathcal{E}(f_j)\}
$$\normalsize
with a probability of at least $\left(1-\frac{K+1}{\sqrt{N}} \right)^H$.
\end{theorem}

\section{Composite Network Construction}
The theoretical analysis in the previous section suggests that with high probability, a trained composite network  performs better than any of its pre-trained components. It also encourages users to apply their domain expertise to design and train critical pre-trained components and incorporate them in their composite network. In this section, we propose heuristic algorithms for composite network construction. Ensemble learning is a simple case of the composite network that will be evaluated and compared with the proposed algorithm. 

For a given set of components, we define the component whose output gives an answer to the main problem as a base component. If the outputs of a component do not directly answer the main problem, we call this an auxiliary component. For example, in the problem of PM2.5 value prediction, the base components output their PM2.5 predictions, whereas a component predicting weather conditions such as wind speed and precipitation is categorized as an auxiliary component.

\begin{algorithm}
\footnotesize{
\caption{Deep Binary Composite Network}
\label{Algo_1}
\KwIn{$\mathcal{F}=\{f_j\}_{0}^{K_1}\cup\{f_{W_j}\}_{K_1+1}^{K}$, a set of activation functions $\mathcal{A}$, pruning threshold $\Delta$}
\KwOut{$g_K$}
$g_1\leftarrow f_1$;
$\forall j \leq K, \mathcal{T}_{j}\leftarrow \emptyset $ \\
\For{$j=2$ to $K$}{
\For{$\sigma(\cdot)\in \mathcal{A}$ }{
 \eIf{$j\leq K_1$}{$\mathcal{T}_{j}\leftarrow \mathcal{T}_{j}\cup\{\sigma(g_{j-1},f_{j}) \}$}
 {$\mathcal{T}_{j}\leftarrow \mathcal{T}_{j}\cup\{\sigma(g_{j-1},f_{W_j}) \}$}
 }
 Train all $h\in \mathcal{T}_{j}$ \\
$g_j\leftarrow \mathsf{argmin}_{h\in{\mathcal{T}_{j}}} \{\mathcal{E}(h)\} $ } 
\For{$j=K$ to $2$}{
\eIf{ $\mathcal{E}(g_{j})-\mathcal{E}(g_{j-1}) \leq \Delta$ } {
$g_j\leftarrow g_{j-1}$
} {output $g_j$ \\ break}
}  
}\normalsize
\end{algorithm} 
The Deep Binary Composite Network (DBCN) Algorithm depicted in Algorithm~\ref{Algo_1} is a greedy method,
the main idea of which is to construct a composite network by inserting one component at a time in some particular order. After each insertion, the depth of the network is increased by 1, as described in Lemma~\ref{addDeep}. We consider the base components first in the insertion order since a base component answers the main problem and it makes sense to use auxiliary components to enhance the performance of the base components later. The pre-trained components are considered before the non-instantiated ones, as pre-trained components are commonly well-crafted and performance-proven. Thus, we insert the components such that pre-trained components are ahead of non-instantiated components, and for each pre-trained and non-instantiated set, the base components are ahead of auxiliary components; finally, the components with lower L2 errors are before those with higher L2 errors.

Algorithm~\ref{Algo_1} takes pre-trained components $\{f_j\}_{1}^{K_1}$ and non-instantiated components $\{f_{W_j}\}_{K_1+1}^{K}$, sorted according to the criteria in the previous paragraph, as inputs, and outputs a deep binary composite network. 
Line 1 initializes the variables used in this algorithm. The first-level \textbf{for} block (from Lines 2 to 12) computes the composite network $g_j$ of depth $j$, iteratively. 
The second-level \textbf{for} block from Lines 3 to 9 generates possible composite networks with both linear and modified logistic activation functions  $\sigma(\cdot)$.
In Line 10, we use traditional stochastic gradient descent backpropagation to train every composite network in $\mathcal{T}_{j}$. Line 12 finds the composite network with the lowest L2 error. Lines 13 to 20 prune the obtained $\{g_{j}\}$ to avoid over-fitting. Once the L2 loss gain is larger than a specified pruning threshold $\Delta$, the pruning process stops and the algorithm outputs the current $g_{j}$; otherwise, $g_{j-1}$ is examined as a consequence.

\begin{algorithm}
\footnotesize{
  \caption{Balanced Base Composite Network}
  \label{Algo_2}
  \KwIn{$\mathcal{F}=\{f_j\}_{0}^{K_1}\cup\{f_{W_j}\}_{K_1+1}^{K}$, a set of activation functions $\mathcal{A}$, the number of base components $K_0$, pruning threshold $\Delta$}
  \KwOut{$g_{K}$}
$\forall j\in [K_0], h_{0,j}\leftarrow f_j $\\
$\forall s\leq \lceil \log_2(K_0) \rceil, t \leq \lceil K_0/2^{s} \rceil, {\mathcal{T}_{s,t}}\leftarrow \emptyset$ \\
$\forall j\leq K, {\mathcal{T}_{j}}\leftarrow \emptyset$ \\
\For{$s=1$ to $\lceil \log_2(K_0) \rceil $}{
 \For{$t=1$ to $\lceil K_0/2^{s} \rceil $}{
 \eIf{$\lceil K_0/2^{s-1} \rceil$ \text{is an odd number} \& $t=\lceil K_0/2^{s} \rceil$ }
  {
  $h_{s,t}\leftarrow h_{s-1,2t-1}$ \;
  }
  {
  \For{$\sigma(\cdot)\in \mathcal{A}$}{
  $\mathcal{T}_{s,t}\leftarrow \mathcal{T}_{s,t}\cup\{\sigma(h_{s-1,2t-1},h_{s-1,2t})\}$\; }
  Train all $h\in \mathcal{T}_{s,t}$\;
  $h_{s,t}\leftarrow  \mathsf{argmin}_{h\in{\mathcal{T}_{s,t}}}\{\mathcal{E}(h)\} $ \;
  }
 }
}
$g_{K_0}\leftarrow h_{\lceil \log_2(K_0) \rceil,1}$ \\
Run Algorithm 1 on $\large(\{g_{K_0}\}\cup\mathcal{F}\setminus \{f_j\}_{j\in[K_0]} \large)$
}\normalsize
\end{algorithm}

The second algorithm, Balanced Base Composite Network (BBCN), is presented in Algorithm~\ref{Algo_2}. The first-level \textbf{for} block (from Lines 4 to 16) generates a flat composite network from the base components, in which each iteration constructs a level of the composite network. The \textbf{for} block (Lines 5 to 15) combines a pair of two base components or two subtrees. Line 18 calls Algorithm~\ref{Algo_1} to complete the execution. In general, Algorithm~\ref{Algo_1} generates a deep binary composite network, whereas Algorithm~\ref{Algo_2} constructs a more balanced composite network, as shown in Fig.~\ref{24hrOfAlgo1And2}. 

\section{PM2.5 Predictions}
In this section, we design five pre-trained components and a non-instantiated component and apply composite network construction methods including exhaustive search, ensemble learning~\cite{zhou2012ensemble}, and Algorithm 1 (DBCN) and Algorithm 2 (BBCN) for PM2.5 prediction. Real-world open data was used to numerically compare the performance of different construction methods and to examine the correctness and efficacy of the proposed theory. In addition, we also compared the methods with traditional machine learning methods, namely, SVM~\cite{Hearst:1998:SVM:630302.630387} and random forests~\cite{Breiman:2001:RF:570181.570182}. For the hardware and software environment,  each of the three servers used in this evaluation was equipped with two Intel Xeon CPUs, 128GB memory, four NVIDIA 1080 GPUs, the Linux operating system, and Keras and Tensorflow as deep learning platforms. 

\subsection{Datasets}
The open data were from two sources: the Environmental Protection Administration (EPA) for air quality data~\cite{EPAdata}, and the Central Weather Bureau (CWB) for weather data~\cite{CWBdata}. There are 21 features in the EPA dataset including values such as PM2.5, PM10, SO$_2$, CO, NO, and NO$_{x}$. The EPA air quality data were collected from eighteen monitoring stations recorded hourly. The second dataset, the CWB open data, has one record per six hours, collected from 31 monitoring stations with 26 features, including temperature, dew point, precipitation, and wind speed and direction. In this study, for all evaluations, the data of years 2014 and 2015 were used as training data and those of 2016 as testing data.

We created a grid of $30\times 38=1140 \;\text{km}^2$ covering the Taipei area, each block of which was $1 \times 1 \;\text{km}^2$. The EPA and CWB data were loaded into the corresponding blocks so that both datasets were temporally aligned at the hour scale (i.e., one record per hour). Interpolation was applied to the CWB data to downscale from 6 hours to 1 hour.  
Note that there were $1140$ blocks in the grid, whereas there were only $18$ EPA stations and $31$ CWB stations; thus more than 1000 blocks were empty, i.e. without EPA or CWB data. 
We adopted the KNN method ($K=4$, i.e., averaging the values of the four nearest neighbors) to initialize the values of the empty blocks, as discussed in \cite{wong2004comparison}.

\subsection{Pre-trained Component Design}
Here we introduce the design rationales of the five pre-trained components in this evaluation. 
As PM2.5 dispersion is highly spatially and temporally dependent, we designed four pre-trained components as base components to model this dependency. Among these, two were convolutional LSTM neural networks (ConvLSTMs~\cite{xingjian2015convolutional}) with the EPA data (denoted as $f_1$) and CWB data (denoted as $f_2$) as input; the other two were fully connected neural networks (FNNs) with the EPA data (denoted as $f_3$) and CWB data (denoted as $f_4$) as input. To model the temporal relationship conveniently using the neural network, the data was fed to the pre-trained components one sequence at a time. We used two pairs of components---$f_1$ and $f_2$, and $f_3$ and 
$f_4$---for the same functions to determine whether component redundancy improves performance.  The fifth pre-trained component (denoted as $f_5$) was to model the association between time and the PM2.5 value.

\begin{table}[ht]
\centering \tiny{
  \caption{LSTM (LsM) v.s. ConvLSTM (CvL)}
  \label{Table_LSTMpkCvLSTM}

  \begin{tabular}{llrr|rr}  
\toprule
Hour& Dataset  & \multicolumn{2}{c}{EPA} & \multicolumn{2}{c}{CWB}\\ \midrule 
&Models & {Training} & {Testing} & {Training} & {Testing}  \\
\midrule
+24h &LsM &8.4158 &10.9586 &8.2741 &11.3947 \\
    &CvL &7.5873 & \textbf{10.5789} & 8.5529 & \textbf{11.2074} \\
\midrule
+48h &LsM &8.7185 &11.5229 &8.5232 &11.8144 \\
    &CvL &8.6541 & \textbf{11.3904} &8.2890 & \textbf{11.7081} \\
\midrule
+72h &LsM &8.7530  &11.7329 & 8.8905 &11.8672 \\
     &CvL &8.8170 & \textbf{11.5279} & 9.2177 & \textbf{11.7756} \\
\bottomrule

\end{tabular} }
\end{table} 

The first experiment was designed to examine the effect of the grid structure in capturing the spatial relationship by comparing the outcomes of LSTM and ConvLSTM. The LSTM model only used the EPA and CWB data without spatial information about the monitoring stations, whereas the ConvLSTM model used the grid data (i.e., considering the whole 1140 blocks with KNN ($K=4$) initialization). 
The accuracy of both models measured in RMSE is presented in Table~\ref{Table_LSTMpkCvLSTM}, which shows the ConvLSTM performs consistently better for the +24h (next 24 hours), +48h (next 48 hours), and +72h (next 72 hours) predictions. Hence, we selected ConvLSTM as the model for $f_1$ and $f_2$.

\begin{table}[ht]
  \centering \tiny{
  \caption{Various configurations of pre-trained components}
  \label{Table_DepthOptimal}
\begin{tabular}{lr|rr|rr|rr}  
\toprule
& {Forecast}   & \multicolumn{2}{c}{+24h}   & \multicolumn{2}{c}{+48h}  & \multicolumn{2}{c}{+72h} \\
               \cmidrule(r){3-4} \cmidrule(r){5-6}\cmidrule(r){7-8}  
Model & {Train.Params} &{Training} & {Testing}  & {Training} & {Testing} &  {Training} & {Testing}  \\
\midrule
$f_{1}$  & 917492 & 7.5873 & \textbf{10.5789}  &8.6541 & \textbf{11.3904}  & 8.8170 & \textbf{11.5279} \\ 
$f_{1,Wr}$  & 3632482 & 9.3054 & 11.9440   &9.1503 & 11.6550  & 8.1616 & 11.7556  \\ 
$f_{1,Dr}$  & 1278692 & 7.6342 & 10.9471   &8.6297 & 11.4844  &9.0803 & 11.5993 \\ 
\midrule
$f_{2}$  & 916908 & 8.5529 & \textbf{11.2074}   & 8.2890 & \textbf{11.7081}  & 9.2177 & \textbf{11.7756} \\ 
$f_{2,Wr}$  & 3631322 & 7.0685 & 11.4974   & 9.2233 & 12.0710  & 9.1766 & 11.9827 \\ 
$f_{2,Dr}$  & 790828 & 6.5404 & 11.7970   &8.4491 & 8.4491  & 9.1500 & 11.9162 \\ 
\midrule
$f_{3,(2)}$  & 1038054  &11.6064 & \textbf{10.8907} & 11.9008 & \textbf{11.6977}   &12.1729 & 11.9999  \\  
$f_{3,(3)}$  & 1068538  & 11.5648 &10.9179  & 11.9726 & 11.7017  & 12.0585 & \textbf{11.9414} \\ 
\midrule
$f_{4,(2)}$  & 582038 & 11.8238 & 11.3400 & 11.6948 & \textbf{11.6147}  & 11.9484 &11.8687 \\  
$f_{4,(3)}$  & 603318 & 11.8253 & \textbf{11.2748}  & 11.7112 & 11.6176  & 12.0199 &\textbf{11.7512} \\ 
\bottomrule
\end{tabular} }
\end{table} 

In the second experiment, we trained the four pre-trained components ($f_1$, $f_2$, $f_3$, $f_4$) individually with different configurations. For instance, we trained the ConvLSTM models ($f_1$ and $f_2$) with a normal configuration, a deeper one (denoted as $Dr$) with stack of two LSTMs, and a wider one (denoted as $Wr$) with a double-width ConvLSTM. Similarly, FNN models $f_3$ and $f_4$ were trained with two or three hidden layers, (denoted as $f_i,{(2 or 3)}$). Their performance was measured by RMSE as shown in Table~\ref{Table_DepthOptimal}. The best performing configurations were selected for the pre-trained components in the following experiments.   

Note that instead of using execution time as a measurement of time complexity, we indicated the complexity using the number of trainable (tunable)  parameters in our study, as shown in the second column of Table~\ref{Table_DepthOptimal}, as the execution times varied widely even for the same training configuration due to diverse server execution contexts, randomness incurred from training commands, and hyperparameter tuning setups. 

\begin{figure}
\centering{
\includegraphics[width=0.46\textwidth]{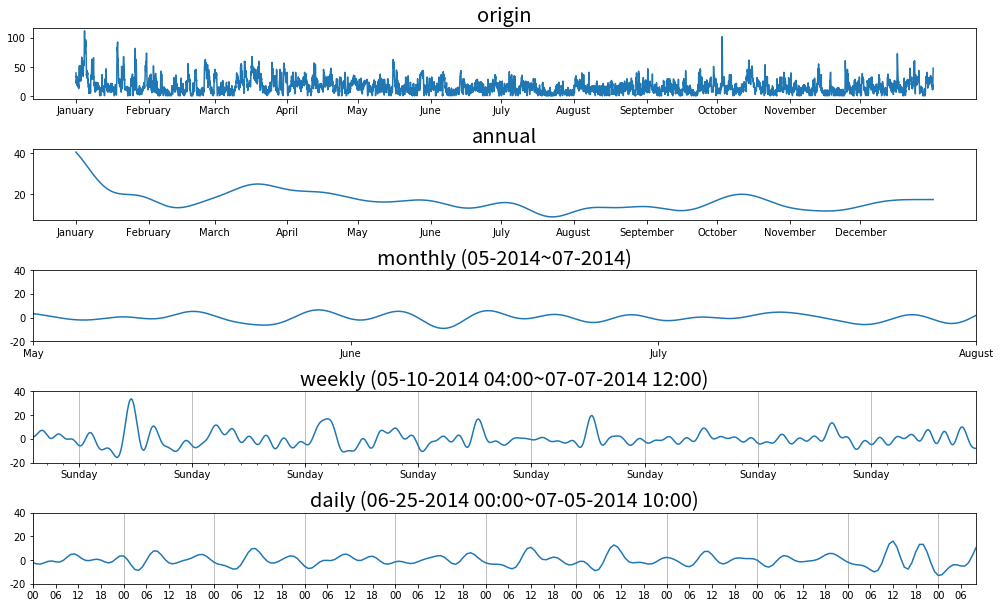}
}
\caption{Annual PM2.5 values at different frequencies} 
\label{WhyTime}
\end{figure}
The fifth pre-trained component ($f_5$) is the association between time and PM2.5 value, which is highly temporally dependent. 
 
Fig.~\ref{WhyTime} shows the PM2.5 values resulting from the different frequency filters~\cite{akansu2001signal}. The top figure shows the original PM2.5 values of the Taitung EPA station in 2014 and the second figure shows the annual trend, which clearly shows that cold months are prone to high PM2.5 pollution. The third graph shows the PM2.5 trend from May to July, which does not reveal a consistent pattern. The fourth figure shows the trends within a week: we observe lower PM2.5 values during the weekend. The fifth figure is the daily trend: PM2.5 values are lower after midnight. Based on these observations, we generated an embedding~\cite{Bengio:2003:NPL:944919.944966} of features including the month, day of the week, and the hour of the day, and trained a LSTM model labeled with PM2.5 values as the pre-trained component $f_5$.

\subsection{Composite Network }
\begin{table}[ht]
\tiny{
\caption{Pre-trained components and testing RMSE}
  \label{Table_f1f2f3f4f5}
\begin{tabular}{llrrr}  
\toprule
Component & Data    & +24h   &+48h  & +72h \\
\midrule
$f_1$: ConvLSTM (2 CNN layers, 1 LSTM) & EPA& 10.5789   & 11.3904  & 11.5279 \\  
$f_2$: ConvLSTM (2 CNN layers, 1 LSTM) & CWB& 11.2074  & 11.7081  & 11.7756  \\  
$f_3$: FNN (2 hidden layers) & EPA & 10.6459  & 11.3291  & 11.6169 \\  
$f_4$: FNN (2 hidden layers) & CWB & 11.5112 &  11.6915  &  11.8017 \\  
$f_5$: LSTM & hr-week-month & 11.4738  &  11.5359 &  11.4540 \\  
\midrule
\multicolumn{5}{l}{EPA 9 features: CO, NO, NO2, NOx, O3, PM10, PM2.5, SO2, THC } \\
\multicolumn{5}{l}{CWB 5 features: AMB-TEMP, RH, rainfall, wind direction-speed (represented as a vector) }
\\
\bottomrule
\end{tabular} }
\end{table} 

There are five pre-trained components from $f_1$ to $f_5$ and one non-instantiated auxiliary component, denoted as $f_{W_6}$, for the composite network construction.  The model of $f_{W_6}$ is a convolutional neural network (CNN) with CWB weather data and forecasts as input to predict upcoming precipitation. The six components are connected by activation functions, either a linear function or a scaled logistic function  ($S(z)=2000/(1+e^{-z/500})-1000$). Note that any activation function that meets all six assumptions in Sec.~3 could be used; for simplicity, we used only the scaled logistic function. 
The prediction accuracy in RMSE of all five pre-trained components is listed in Table~\ref{Table_f1f2f3f4f5}. Note that in this study we did
not set out to design an optimized composite network for the best PM2.5 prediction. Rather, our main purpose was to implement and evaluate the proposed composite network theory. Nevertheless, the design of components and composite network follows the advice of domain experts and exhibits reasonably good performance in PM2.5 prediction.

\subsubsection{DBCN and BBCN}

The step-by-step running of Algorithm~\ref{Algo_1} (DBCN) and the results are shown in Table~\ref{Table_Algo1_24hr} for the +24h predictions. First, $f_1$ is automatically selected as $g_1$, after which $f_3$ is included, as it has the lowest RMSE among the remaining components. In the first column of the table, $L(g_1,f_3)$ has a lower RMSE than $SL(g_1,f_3)$ and is selected as $g_2$, as marked in 
the last column (``Front-runner'').   
(Note that $SL$ is an abbreviation of the scaled logistic function cascading a linear function.) Next, Algorithm~1 generates the composite network $L(g_5,f_{W_6})$ with a testing RMSE of 10.9531 for the +24h prediction. 
Table~\ref{Table_Algo1_48_72hr} shows the +48h and +72h prediction results: 
the generated models are different from each other and the model for +24h.  

The ``Trainable/total'' column indicates the number of trainable parameters and total parameters during the training phase. The trainable parameters are updated during each backpropagation stochastic gradient descent optimization, and the total parameters are the number of trainable parameters plus the fixed parameters in the pre-trained components. As only the trainable parameters are updated during training, the composite network framework may greatly alleviate many burdens in training a complicated composite network.

\begin{table}[ht]
\centering \tiny{ 
\caption{Composite networks using Algo 1: DBCN, +24h} \label{Table_Algo1_24hr}
\begin{tabular}{l|rrr|l}  
\toprule
      &\multicolumn{2}{c}{RMSE}  & Parameters & \\
     \cmidrule(r){2-3}
Model &Training & \textbf{Testing} &Trainable/total & Front-runner\\
\midrule
$g_1\leftarrow f_1$& & & 0/ &  \\
$L(g_1,f_3)$  & 7.2128 & \textbf{10.2277} &666/1956230 & $g_2$\\  
$SL(g_1,f_3)$  & 7.3311 & 10.3454  &666/1956230  &\\ 
\midrule
$L(g_2,f_2)$  & 7.1364 & 10.2410   &666/2873814 & \\  
$SL(g_2,f_2)$  & 7.3208 & \textbf{10.2409}  & 666/2873814&$g_3$ \\ 
\midrule
$L(g_3,f_4)$  & 7.0787 & \textbf{10.3039}  &666/3456518 & $g_4$\\  
$SL(g_3,f_4)$  & 7.1931 & 10.3501  & 666/3456518 & \\ 
\midrule
$L(g_4,f_5)$  & 7.0911 & 10.3275 &666/4411100 & \\  
$SL(g_4,f_5)$  & 7.0560 & \textbf{10.2119} &666/4411100& $g_5$\\ 
\midrule
$L(g_5,f_{W_6})$  & 6.9608 & 10.1131  & 42046/4453146 & \\  
$SL(g_5,f_{W_6})$  & 6.9705 & \textbf{10.1053} &42046/4453146& $g_6$\\ 
\bottomrule
\end{tabular} }
\end{table} \normalsize 

\begin{table}[ht]
\centering \tiny{
\caption{Composite networks using Algo 1: DBCN, +48h, +72h} \label{Table_Algo1_48_72hr}
\begin{tabular}{ll|rrr|l}  
\toprule
    &  &\multicolumn{2}{c}{RMSE}  & Parameters & \\
     \cmidrule(r){3-4}
Prediction & Model &Training & \textbf{Testing} &Trainable/total & Front-runner\\
\midrule
+48h & $SL(g_4,f_2)$  & 8.0678 & \textbf{11.0469}&666/4411100& $g_5$\\ 
\midrule
& $SL(g_5,f_{W_6})$  &7.8941 & \textbf{10.9531}   &42046/4453146 & $g_6$\\  
\midrule
\midrule
+72h & $L(g_4,f_3)$  & 8.2305 & \textbf{11.4274}    &666/4411100& $g_5$\\  
\midrule
& $L(g_5,f_{W_6})$  & 8.2448 & \textbf{11.2541} & 42046/4453146 & $g_6$\\  
\bottomrule
\end{tabular} }
\end{table} \normalsize 

\begin{table}[ht]
\centering \tiny{
\caption{Composite networks using Algo 2: BBCN, +24h} \label{Table_Algo2_24hr}
\begin{tabular}{l|rrr|l}  
\toprule
      &\multicolumn{2}{c}{RMSE}& Parameters & \\
      \cmidrule(r){2-3}
Model &Training & \textbf{Testing} &Trainable/total & Front-runner\\
\midrule
\multicolumn{5}{l}{$h_{0,1}\leftarrow f_1,h_{0,2}\leftarrow f_2$}  \\
$L(h_{0,1},h_{0,2})$  & 7.1016 & \textbf{10.4075}  & 666/1835094 & $h_{1,1}$\\  
$SL(h_{0,1},h_{0,2})$  & 6.5801 & 10.4581   &666/1835094 &\\ 
\midrule
\multicolumn{5}{l}{$h_{0,3}\leftarrow f_3,h_{0,4}\leftarrow f_4$} \\
$L(h_{0,3},h_{0,4})$  & 11.4359 & \textbf{10.7670} & 666/1620758& $h_{1,2}$\\  
$SL(h_{0,3},h_{0,4})$  & 11.5389 & 10.8508  &666/1620758 &\\ 
\midrule
$L(h_{1,1},h_{1,2})$  & 7.2375 & 10.4536  & 666/3456518 & \\  
$SL(h_{1,1},h_{1,2})$  & 7.2523 & \textbf{10.3226} &666/3456518 & $h_{2,1}$ \\ 
\midrule
\multicolumn{5}{l}{$h_{2,2}\leftarrow h_{1,3}\leftarrow h_{0,5}\leftarrow f_5$} \\
$L(h_{2,1},h_{2,2})$  & 7.1069 & \textbf{10.4712}  & 666/4411100 & $h_{3,1}$\\  
$SL(h_{2,1},h_{2,2})$  & 7.1202 & 10.5064  &666/4411100 &\\ 
\midrule
\multicolumn{5}{l}{$g_5\leftarrow h_{3,1}$} \\
$L(g_5,f_{W_6})$  & 6.9828 & \textbf{10.1938} &  42046/4453146 & $g_6$\\  
$SL(g_5,f_{W_6})$  & 6.9964 & 10.2257  & 42046/4453146&\\ 
\bottomrule
\end{tabular} }
\end{table} \normalsize 

\begin{table}[ht]
\centering \tiny{
\caption{Composite networks using Algo 2: BBCN, +48h, +72h} \label{Table_Algo2_48_72hr}
\begin{tabular}{ll|rrr|l}  
\toprule
    &  &\multicolumn{2}{c}{RMSE}  & Parameters & \\
     \cmidrule(r){3-4}
Prediction & Model &Training & \textbf{Testing} &Trainable/total & Front-runner\\
\midrule
+48h & $SL(h_{2,1},h_{2,2})$  & 7.9949 & \textbf{11.0516} &666/4411100& $h_{3,1}$ \\ 
\midrule
& $L(g_5,f_{W_6})$  & 8.5736 & \textbf{11.0182}  &42046/4453146 & $g_6$\\  
\midrule
\midrule
+72h & $L(h_{2,1},h_{2,2})$  & 8.4460 & \textbf{11.5100}  & 666/4411100 & $h_{3,1}$\\  
\midrule
& $L(g_5,f_{W_6})$  & 9.1848 & \textbf{11.4153}  & 42046/4453146& $g_6$\\  
\bottomrule
\end{tabular} }
\end{table} \normalsize 

\begin{figure}
\centering{
\includegraphics[width=0.45\textwidth]{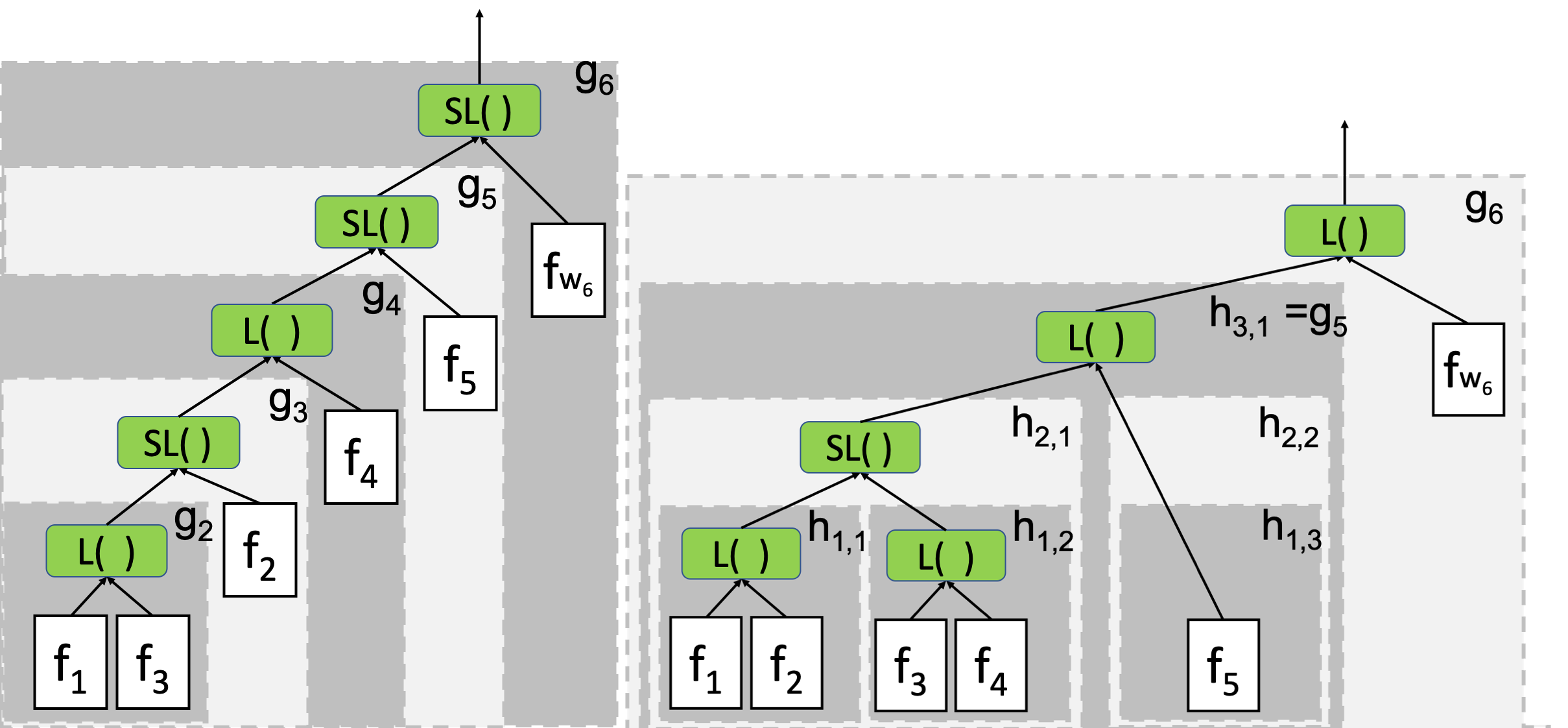}
}
\caption{Composite networks using Algorithms~\ref{Algo_1} (left) and \ref{Algo_2} (right) for +24h prediction} 
\label{24hrOfAlgo1And2}  
\end{figure}

The processes and results of Algorithm~\ref{Algo_2} (BBCN) are shown in Table~\ref{Table_Algo2_24hr} for the +24h PM2.5 predictions and in Table~\ref{Table_Algo2_48_72hr} +48h and +72h. Note that Algorithm 2 constructs a composite network by merging the base components in the beginning: the first row of Table~\ref{Table_Algo2_24hr} combines $f_1$ and $f_2$, and the second row combines $f_3$ and $f_4$. Generally, both DBCN and BBCN methods meet the claim of the proposed composite network theory: combining more pre-trained components yields improved RMSE results. The composite networks constructed using Algorithms~\ref{Algo_1} and \ref{Algo_2} for +24h prediction are contrasted in Fig.~\ref{24hrOfAlgo1And2}.

\subsubsection{Exhaustive Search Construction}

\begin{table}[ht]
\centering \tiny{
\caption{Composite networks of $f_1,f_2$ using exhaustive search, +24h}\label{Table_Compositef1f2}
\begin{tabular}{r|lrrr}  
\toprule
Step & Model &{Training} & {Testing} &  Front-runner  \\
\midrule
1&$L(f_1^{\times},f_2^{\times})$ & 7.1016 & 10.4075 & \\
&$L(f_1^{\times},f_2^{\circ})$ & 6.5417 & 10.2097 &  \\
&$L(f_1^{\circ},f_2^{\times})$ & 6.6574 & 10.3484 &   \\
&$L(f_1^{\circ},f_2^{\circ})$ & 6.3394 & 10.0423 &    \\
&$SL(f_1^{\times},f_2^{\times})$ & 6.5801 & 10.4581 &  \\
&$SL(f_1^{\times},f_2^{\circ})$ & 6.6648 & 9.9048 &  \\
&$SL(f_1^{\circ},f_2^{\times})$ & 6.5052 & 10.1654 &  \\
&$SL(f_1^{\circ},f_2^{\circ})$ & 6.5109 & \textbf{9.7275} &  $g_1$  \\
\bottomrule
\end{tabular} }
\end{table} 

\begin{table}[ht]
\centering \tiny{
\caption{Composite networks of  $f_3,f_4$ using exhaustive search, +24h}
  \label{Table_Compositef3f4}
\begin{tabular}{r|lrrr}  
\toprule
Step & Model &{Training} & {Testing} &  Front-runner  \\
\midrule
2&$L(f_3^{\times},f_4^{\times})$ & 11.4359 & 10.7670 &  \\
&$L(f_3^{\times},f_4^{\circ})$ & 10.9690 & 10.8618 & \\
&$L(f_3^{\circ},f_4^{\times})$ & 11.0442 & 10.8285 &   \\
&$L(f_3^{\circ},f_4^{\circ})$ & 11.2553 & 10.7017 &   \\
&$SL(f_3^{\times},f_4^{\times})$ & 11.5389 & 10.8508 &    \\
&$SL(f_3^{\times},f_4^{\circ})$ & 11.2916 & 10.9000 & \\
&$SL(f_3^{\circ},f_4^{\times})$ & 11.1600 & 10.8505 &   \\
&$SL(f_3^{\circ},f_4^{\circ})$ & 11.1543 & \textbf{10.6877} & $g_2$ \\
\bottomrule
\end{tabular} }
\end{table} 

In this subsection, an exhaustive search method based on Algorithm 2 is introduced to construct a high-accuracy PM2.5 prediction composite network for use as a high-mark benchmark for comparison. In contrast to the previous approaches, in the exhaustive search approach the parameters inside a pre-trained component can be either fixed or open in order to guarantee the best construction. Hence, instead of the 5 pre-trained and 1 non-instantiated components used by the previous algorithms, we now have five additional pre-trained components with open (tunable) parameters (i.e., non-instantiated components). The new notation $^{\times}$ denotes pre-trained components and $^{\circ}$ denotes non-instantiated components. For instance, $f_{1}^{\circ}$ is component 1 but non-instantiated. Inherently, with exhaustive search the construction takes a substantially longer time to complete (i.e., with time complexity of $O(2^K)$ ), but has the potential for better performance. A complete exhaustive search example for PM2.5 prediction is conducted to evaluate the performance improvement.

For +24h prediction, the the exhaustive search algorithm employs the same composite network layout as Algorithm 2. The best composition combining $f_1$ and $f_2$ is $g_1 = SL(f_1^{\circ},f_2^{\circ}))$, as shown in Table~\ref{Table_Compositef1f2}, which corresponds to combining non-instantiated $f_1$ and $f_2$ and applying the scaled logistic activation function results in the lowest RMSE. In the next step, $f_3$ and $f_4$ are combined with the front-runner as $g_2 = SL(f_3^{\circ},f_4^{\circ}))$ as shown in Table~\ref{Table_Compositef3f4}. Step 3 considers all possible combinations of $g_1$ and $g_2$ to find the best $g_3$, as shown in Table~\ref{compositeG1G2}. Note that we treat $g_i^{\circ}$ as having all non-instantiated components; 
for $g_i^{\times}$, all components are pre-trained.   

\begin{table}[ht]
\centering \tiny{
\caption{Composite networks of $g_1,g_2$ using exhaustive search, +24h}
  \label{compositeG1G2}
\begin{tabular}{r|lrrr}  
\toprule
Step & Model &{Training} & {Testing} &  Front-runner  \\
\midrule
3&$L(g_1^{\times},g_2^{\times})$ & 5.8897 & 9.6059 & \\
&$L(g_1^{\times},g_2^{\circ})$ & 5.6321 & 9.5278 &  \\
&$L(g_1^{\circ},g_2^{\times})$ & 4.9207 & 9.8139 &  \\
&$L(g_1^{\circ},g_2^{\circ})$ & 4.5724 & 9.5881 &    \\
&$SL(g_1^{\times},g_2^{\times})$ & 5.8941 & 9.6162 &   \\
&$SL(g_1^{\times},g_2^{\circ})$ & 5.3703 & \textbf{9.5250} &$g_3$   \\
&$SL(g_1^{\circ},g_2^{\times})$ & 4.7438 & 9.8185 &  \\
&$SL(g_1^{\circ},g_2^{\circ})$ & 4.1957 & 9.6039 &  \\
\bottomrule
\end{tabular} }
\end{table} 

\begin{table}[ht]\centering \tiny{
\caption{Composite networks of $g_3$ and $f_5$, then $f_{W_6}$, +24h}
  \label{Table_G3TimeWeather}
\begin{tabular}{r|lrrr}  
\toprule
Step & Model &{Training} & {Testing} &  Front-runner  \\
\midrule
4a&$L(g_3^{\times},f_5^{\times})$ & 5.3349 & 9.3055 &   \\
&$L(g_3^{\times},f_5^{\circ})$ & 5.2516 & 9.3186 &  \\
&$L(g_3^{\circ},f_5^{\times})$ & 5.5953 & 9.5570& \\
&$L(g_3^{\circ},f_5^{\circ})$ & 6.6938 & 9.4190 &\\
&$SL(g_3^{\times},f_5^{\times})$ & 5.5415 & 9.4504 &  \\
&$SL(g_3^{\times},f_5^{\circ})$ & 5.1646 & \textbf{9.2438} &  $g_{4a}$\\
&$SL(g_3^{\circ},f_5^{\times})$ & 7.2401 & 9.4492&     \\
&$SL(g_3^{\circ},f_5^{\circ})$ & 7.0476 & 9.4947 &  \\
\midrule
5a&$L(g_{4a}^{\times},f_{W_6}^{\circ})$ & 5.3665 & 9.2730 &  \\
&$L(g_{4a}^{\circ},f_{W_6}^{\circ})$ & 4.3968 & 9.4362 &   \\
&$SL(g_{4a}^{\times},f_{W_6}^{\circ})$ & 5.5986 & \textbf{9.1971} &  $g_{5a}$  \\
&$SL(g_{4a}^{\circ},f_{W_6}^{\circ})$ & 5.5421 & 9.4882 &  \\
\bottomrule
\end{tabular} }
\end{table} 
\begin{table}[ht]\centering \tiny{
\caption{Composite networks of $g_3$ and $f_{W_6}$, then $f_5$, +24h}
  \label{composeG3WeatherTime}
\begin{tabular}{r|lrrr}  
\toprule
Step & Model &{Training} & {Testing} &  Front-runner  \\
\midrule
4b&$L(g_3^{\times},f_{W_6}^{\circ})$ & 4.5310 & 9.4551 &  \\
&$L(g_3^{\circ},f_{W_6}^{\circ})$ & 4.6822 & 9.4677&   \\
&$SL(g_3^{\times},f_{W_6}^{\circ})$ & 5.5339 & \textbf{9.3423} & $g_{4b}$ \\
&$SL(g_3^{\circ},f_{W_6}^{\circ})$ & 4.7831 & 9.4991 & \\
\midrule
5b&$L(g_{4b}^{\times},f_5^{\times})$ & 5.6073 & 9.3961&  \\
&$L(g_{4b}^{\circ},f_5^{\times})$ & 6.6991 & \textbf{9.2591} &$g_{5b}$ \\
&$L(g_{4b}^{\times},f_5^{\circ})$ & 5.3298 & 9.2721 &  \\
&$L(g_{4b}^{\circ},f_5^{\circ})$ & 7.1666 & 9.5607&  \\
&$SL(g_{4b}^{\times},f_5^{\times})$ & 5.4710 & 9.3313 & \\
&$SL(g_{4b}^{\times},f_5^{\circ})$ & 5.3607 & 9.3130 &   \\
&$SL(g_{4b}^{\circ},f_5^{\times})$ & 6.2875 & 9.3586 &   \\
&$SL(g_{4b}^{\circ},f_5^{\circ})$ & 6.5281 & 9.5541&  \\
\bottomrule
\end{tabular} }
\end{table} 

\begin{figure}
\centering{
\includegraphics[width=0.45\textwidth]{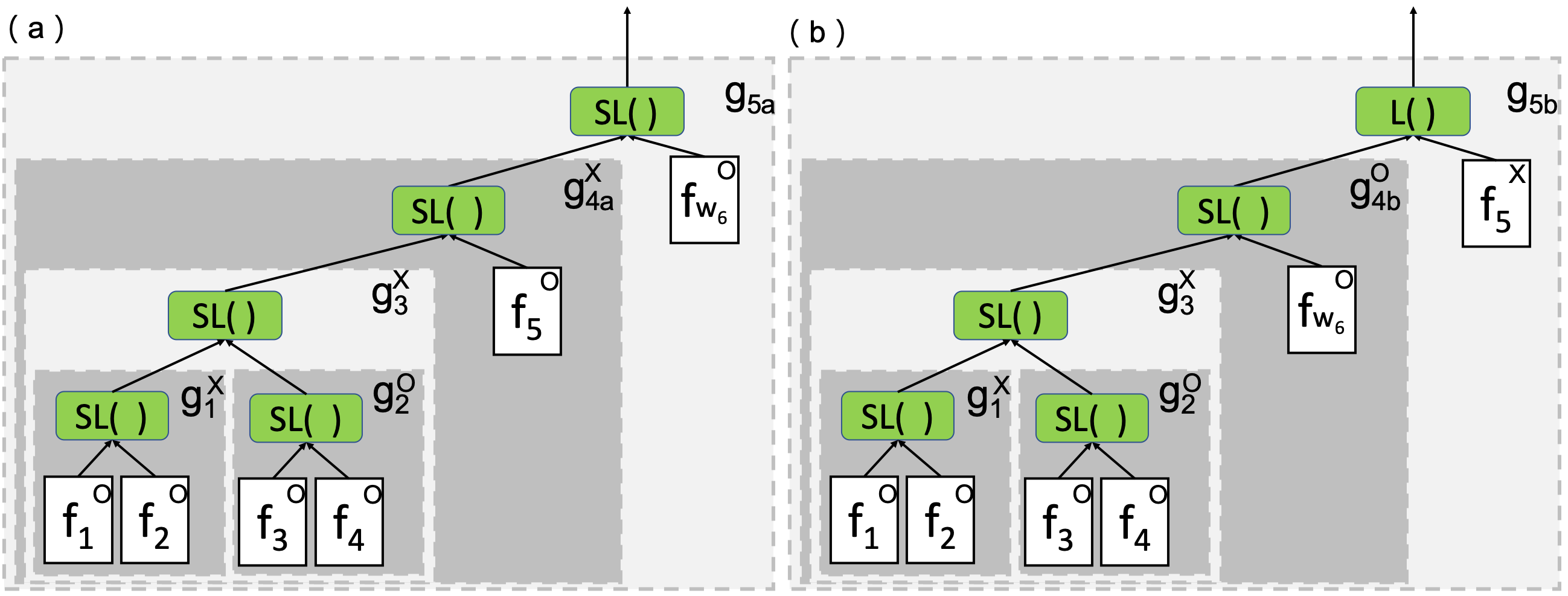}
}
\caption{Composite networks of (a) Tables~\ref{Table_G3TimeWeather} and (b) \ref{composeG3WeatherTime} for +24h prediction} 
\label{IllusionG5a}
\end{figure}

Now only $f_5$ and $f_{W_6}$ are not combined. Here we examine different sequences of $f_5$ and $f_{W_6}$. In Steps 4a and 5a, $f_5$ is considered first and then $f_{W_6}$. The results are shown in Table~\ref{Table_G3TimeWeather}. Steps 4b and 5b consider the opposite sequence from the results listed in Table~\ref{composeG3WeatherTime}.  The best models of $g_{5a}$ and $g_{5b}$ are illustrated in Fig.~\ref{IllusionG5a}. The composite networks for +48h and +72h predictions using exhaustive search were conducted accordingly and their results are used for performance comparisons in the next subsection.

\subsubsection{ Comparisons of All Methods}

\begin{table}[ht]
\centering 
\tiny{
\caption{Summary of all methods (RMSE)}
  \label{Summary_all_methods}
\begin{tabular}{lr|rr|rr|rr}  
\toprule
&{}   & \multicolumn{2}{l}{+24h}   & \multicolumn{2}{l}{+48h}  & \multicolumn{2}{l}{+72h} \\
               \cmidrule(r){3-4} \cmidrule(r){5-6}\cmidrule(r){7-8}  
Method & {Trainable} &{Training} &{Testing}  & {Training} & {Testing} &  {Training} & {Testing}  \\
\midrule
SVM & - &  11.6440 & 10.9117 & 12.1246 & 11.5469 & 12.1670 &  11.6376     \\ 
Random forests & - & 3.3181  & 10.9386  & 3.4304 & 11.9037 &  3.4148 &  12.0917  \\ 
Ensemble & 1638 & 11.6955  & 11.0200  & 12.2609 & 11.3969 &  12.6605 &  11.6119  \\ 
\emph{SL}(Ensemble) & 1638  & 11.5855 & 10.9184 & 12.2080 & 11.2815 & 12.5690 &  11.5411    \\ 
DBCN$_{ \mathsf{Relu}}$ & 2664  & 12.4800 & 11.4540  & 13.3464 & 12.1947 &  14.0421 &  12.6546    \\ 
DBCN$_{ \mathsf{Sigm}}$ & 4032  & 11.7786 & 10.9803  & 13.6521 & 12.4418 &  13.4414 &  12.2825    \\
DBCN  & 2664  & 7.0560 & \textbf{10.2119}  & 8.0678 & \textbf{11.0469} &  8.2305 &  \textbf{11.4274}    \\ 
BBCN$_{\mathsf{Relu}}$   & 2664 & 13.3711 & 12.4575  & 14.6168 & 13.2662 &  15.8200 &  14.0754 \\ 
BBCN$_{\mathsf{Sigm}}$  &  4032 & 12.5376 & 11.4600  & 13.0951 & 12.2047 &  13.5416 & 12.0388 \\ 
BBCN  &  2664 & 7.1069 & 10.4712  & 7.9949 & 11.0935 &  8.4460 &  11.5100 \\ 
Exhaustive-a  & 2664+$\alpha$  & 5.1646 & \textbf{9.2438}  & 5.0981 & \textbf{10.2402} &  6.7830 &  \textbf{10.4265}    \\ 
\midrule
\midrule
\multicolumn{8}{l}{(Include $f_{W_6}$), note that $\alpha=$4408436, $\beta=$4449816} \\ 
Ensemble  & 43684  & 11.5253 & 10.7338  & 12.4490 & 11.1874 &  12.5822 &  11.4804   \\
\emph{SL}(Ensemble)  & 43684  & 11.5117 & 10.8125  & 12.3939 & 11.1628 &  12.7025 &  11.3376   \\
DBCN$_{\mathsf{Relu}}$ & 44710  & 12.9434 & 11.8209  & 14.3413 & 12.8331 &  14.3562 & 12.7689    \\ 
DBCN$_{\mathsf{Sigm}}$  & 46420 & 11.9444 & 10.9167 & 12.1700 & \textbf{10.9474} &  13.2754 &  11.8630   \\ 
DBCN  &  44710 & 6.9705 & \textbf{10.1053}  & 7.8941 & 10.9531 &  8.2448 &  \textbf{11.2541}   \\
BBCN$_{\mathsf{Relu}}$   &  44710 & 11.4985 & 10.5742  & 12.0386 & 11.0392 &  12.7188 &  11.4047 \\ 
BBCN$_{\mathsf{Sigm}}$  & 46420 & 12.4675 & 11.3664  & 13.1786 & 11.9285 &  13.3815 & 11.8680 \\ 
BBCN &  44710  & 6.9828 & 10.1938  & 8.5736 & 11.0182 &  9.1848 &  11.4153  \\
Exhaustive-a  & 44710+$\beta$  & 5.5986 & \textbf{9.1971}  & 5.1292 & \textbf{10.2190} &  7.9572 &  \textbf{10.3588}    \\
Exhaustive-b &  44710+$\beta$ & 6.6991 & \textbf{9.2591}  & 5.6125 & \textbf{10.0632} &  5.7376 &  \textbf{10.2671}    \\
\bottomrule
\end{tabular} }
\end{table}

In this section, we compare the performance of different composite network algorithms, including DBCN, BBCN, exhaustive search, and ensemble methods, as well as machine learning methods, SVM and random forest. In addition, we use Relu and logistic activation functions to replace the scaled logistic function in DBCN and BBCN to show the performance differences.  
As claimed, the composite network theory guarantees, with high probability, that the composite network has lower RMSE than any of its components, which is supported by all DBCN, BBCN, exhaustive search, and ensemble methods. 

We summarize the results of all methods in Table~\ref{Summary_all_methods} for RMSE, and in Table~\ref{Summary_all_methods_MAE_SMAPE} for MAE (mean absolute error) and SMAPE (symmetric mean absolute percentage error). For the SVM and random forest experiments, we used the tools from scikit-learn~\cite{scikit} with pre-trained components only (i.e., $f_1$ to $f_5$, with $\alpha$ the total parameters inside these five components.) Likewise with ensemble learning and with ensemble learning with the scaled logistic function as the activation function (denoted as \emph{SL}(ensemble)). The four evaluations yielded close testing RMSE values for all predictions, but the ensemble learning method performed slightly better, while the random forest method seems overfitted, as the training RMSE is low. DBCN performs slightly better than DBNN, and the exhaustive search has the best outcome. For the activation functions, it is interesting to discover that the scaled logistic function performs almost better than the regular logistic and Relu functions.

Now that $f_{W_6}$ is included in composite network construction, it can be seen that DBCN, DBNN, Exhaustive Search (a), and Exhaustive Search (b), as depicted in Fig.~\ref{IllusionG5a}, show improvements over the composite networks without $f_{W_6}$.The sum of parameters inside these six components is denoted as $\beta$ in Table~\ref{Summary_all_methods}. The second column of the table gives the number of trainable parameters for each evaluation; this shows that for the composite network the training parameters are moderate.
Table~\ref{Summary_all_methods_MAE_SMAPE} shows the MAE measurements of the evaluations in Table~\ref{Summary_all_methods}. The ordering of the testing MAE results are very similar to that of the RMSE results.

\section{Related Work}
In this section, we discuss related work in the literature from the perspective of the composite network framework and PM2.5 prediction. For the framework, the composite network is related to the methods such as ensemble learning~\cite{zhou2012ensemble}, transfer learning~\cite{erhan2010does} and model reuse~\cite{yang2017deep,wu2019heterogeneous}. We will also discuss some representative work on air quality prediction.

\textbf{Ensemble Learning.}
Typical ensemble learning methods include bagging, boosting, stacking, and linear combination/regression. 
Since the bagging groups data by sampling and boosting tunes the probability of data~\cite{zhou2002ensembling}, these frameworks are not similar to composite neural networks. However, there are fine research results that are instructive for accuracy improvement~\cite{dvzeroski2004combining,gashler2008decision,zhou2002ensembling}. 
In this work, we consider the neural network composition, but not data enrichment.

Among the ensemble methods, stacking is closely related to our framework. The idea of stacked generalization~\cite{wolpert1992stacked}, in Wolpert's terminology, is to combine two levels of generalizers. The original data are taken by several level-0 generalizers, after which their outputs are concatenated as an input vector to the level-1 generalizer. According to the empirical study of Ting and Witten~\cite{ting1999issues}, the probability distribution of the outputs from level 0, instead of their values, is critical to accuracy. Their experimental results also imply that  multi-linear regression is the best level-1 generalizer, and a non-negative weight restriction is necessary for regression but not for classification.
However, our analysis shows that the activation functions that satisfy Assumption A3 
have a high probability guarantee of reducing the L2 error. In addition, our empirical evaluations show that the scaled logistic activation usually performs well. 

The work of Breiman~\cite{breiman1996stacked} restricts non-negative combination weights to prevent poor generalization errors and concludes that it is not necessary to restrict the sum of weights to equal 1. In \cite{hashem1997optimal}, Hashem shows that linear dependence of components could be, but is not necessarily always, harmful to ensemble accuracy, whereas our work allows a mix of pre-defined and non-instantiated components as well as negative weights to provide flexibility in solution design. 

\textbf{Transfer Learning.} 
In the context of one task with a very small amount of training data with another similar task that has sufficient data, transfer learning can be useful~\cite{pan2009survey}. 
Typically the two data sets---the source and target domains---have different distributions.
A neural network such as an auto-encoder is trained with source-domain data and the corresponding hidden layer weights or output labels are used for the target task. 
Part of transplanted weights can be kept fixed during the consequent steps, whereas others are trainable for fine-tuning~\cite{erhan2010does}. This is in contrast to the composite neural network, in which the pre-trained weights are always fixed. 

\textbf{Model Reuse.}
In recent years, some proposed frameworks emphasize the reuse of fixed models~\cite{yang2017deep, feng2018autoencoder, wu2019heterogeneous}. 
In this framework, pre-trained models are usually connected with the main (i.e., target) model, and then the dependency is gradually weakened by removing or reducing the connections during the training process. In this way, the knowledge of the fixed model is transferred to the main model; the key point is that model reuse is different from transfer learning as well as the composite neural network. 

Pre-trained models are widely applied in applications of natural language processing to improve the generation ability of the main model, such as in BERT~\cite{devlin2018bert} and ELMo~\cite{peters2018deep}.   
Multi-view learning~\cite{zhao2017multi} is another method to improve generalization performance.  In this approach, a specific task owns several sets of features corresponding to different views, just like an object observed from various perspectives, and separate models are trained accordingly. Then, the trained models for different views are combined using co-training, co-regularization, or transfer learning methods. 

\textbf{Air Quality Forecasting.}
There are several air quality prediction systems that combine different components, although these components are usually not pre-trained. In \cite{zheng2013u}, Zheng et al. propose a model  combining two components---an artificial neural network as the spatial classifier and a conditional random field as the temporal classifier---to infer air quality indices. Zheng et al.~\cite{zheng2015forecasting} propose a prediction model for +48h forecasting composed of four components: a temporal predictor (linear regression), a spatial predictor (neural network), a dynamic aggregator of both temporal and spatial predictors, and an inflection predictor capturing sudden changes.
According to the data provided by the monitoring stations, Hsieh et al.~\cite{hsieh2015inferring} propose a system to predict the air quality class even for locations without monitoring stations.
Furthermore, for locations with poor prediction,  a location is recommended to install a new monitoring station for best prediction. Their inference model is based on an affinity graph.
In \cite{wei2016transfer}, Wei et al. employ transfer learning to address the problem of big cities with a large amount of air quality data along with small cities that have insufficient data to train a model from scratch.  Using pre-trained components  shows strengths in flexibility in design and efficiency in training, the work in ~\cite{yi2018deep} presents well-thought component designs, and feature engineering and encoding that are valuable for forthcoming PM2.5 prediction studies. Yi et al.~\cite{yi2020predicting} propose a deep neural network consisting of a spatial transformation component and a deep distributed fusion network to fuse heterogeneous urban data to capture the factors affecting air quality. 
 The hybird architecture of CNN and Bi-LSTM trained from scratch by Du et al. \cite{du2021deep} is designed to learn the correlation and interdependence spatial-temporal information. 
In the reverse of decomposition, Qi et al.\cite{qi2018deep} integrate the three tasks, feature analysis, prediction and interpolation, into one deep learning model.

\section{Conclusions}
In this work, we investigate a composite neural network composed of pre-trained components connected by differentiable activation functions. Through theoretical analysis and empirical evaluations, we show that if assumptions A1 to A4 are satisfied, especially when training data is sufficient, then a composite network has better performance than all of its components with high probability.

While the proposed theory ensures overall performance improvement, it is still not clear how to decompose a complicated problem into components and how to construct them into a composite network to yield acceptable performance. Another problem worth investigating is when the performance improvements diminish even after adding more components. Note that in real-world applications, the amount of data, the data distribution, and the data quality affect performance considerably.

\bibliographystyle{IEEEtran} 
\bibliography{reference.bib} 

\newpage\normalsize
\appendix{}
\textbf{A1. More details of Proofs}\\
\begin{proof} 
(of Lemma \ref{lemma_1}) \\
Recall that in the case of linear activate function, $g(\mathbf{x})= L(f_1,...f_K)=\sum_{j\in[K]^{+}}{ \theta_j}f_j(\mathbf{x})$. Also recall that 
$
\mathcal{E}_{\mathbf{\Theta}}(\mathbf{x};g)
=\sum_{i=1}^{N}{( g(\mathbf{x}^{(i)})-y^{(i)} )^2}.
$ 
To prove the existence of the minimizer, it is sufficient to find the critical point for the deferential of Eq. (1). That is, to calculate the solution, the set of equations: 
$$
\nabla_{ \mathbf{\Theta} } \mathcal{E}\left(\mathbf{x};g\right)
={\left[\begin{array}{c}
        \frac{\partial \mathcal{E}}{\partial\theta_0} \\ 
        \vdots \\
        \frac{\partial \mathcal{E}}{\partial\theta_K} \end{array}\right]} 
={\left[\begin{array}{c}
     0\\
     \vdots \\
     0  \end{array}\right]}, 
$$
where for each $s\in[K]^{+}$, and
$$\aligned
\frac{\partial \mathcal{E}}{\partial\theta_s}
=&2\sum_{i=1}^{N}{\left( g(\mathbf{x}^{(i)})-y^{(i)}  \right)\cdot f_s(\mathbf{x}^{(i)})} \\
=&2\sum_{i=1}^{N}{\left(\sum_{j\in[K]^{+}}{\theta_j}f_j(\mathbf{x}^{(i)}_j)-y^{(i)}  \right)\cdot f_s(\mathbf{x}^{(i)})} \\
=& 2\left(\sum_{j\in [K]^{+}}{\theta_j \langle \vec{f_s},\vec{f_j} \rangle} -\langle \vec{f_s},\vec{y} \rangle \right).
\endaligned$$
Hence, to solve  
$
\nabla_{ \mathbf{\Theta} } \mathcal{E}\left(\mathbf{x};g\right)
={\vec{0}}
$
is equivalent to solve $\theta_t$s in the equation
$$
\left[ \langle \vec{f_s},\vec{f_t} \rangle \right]_{(K+1)\times(K+1)}\times \left[ {\theta}_t\right]_{(K+1)\times 1} = \left[ \langle \vec{f_s},\vec{y} \rangle \right]_{(K+1)\times 1}
$$
where the indexes $s,t$ 
are in $[K]^{+}$.

Note that linear independence of $\{\vec{f}_j\}_{j\in[K]^{+}}$ makes
$\left[ \langle \vec{f}_s,\vec{f}_t\rangle \right]_{(K+1)\times(K+1)}$ a positive-definite Gram matrix \cite{horn1990matrix}, which means the inversion $\left[ \langle \vec{f}_s,\vec{f}_t\rangle \right]_{(K+1)\times(K+1)}^{-1}$ exists.
Then the minimizer $\mathbf{\Theta}^{*}$ is solved:
\footnotesize{
\begin{equation}\label{Formula}
\left[ {\theta}_t\right]_{(K+1)\times 1} 
= 
\left[ \langle \vec{f_s},\vec{f_t} \rangle \right]_{(K+1)\times (K+1)}^{-1}\times
\left[ \langle \vec{f_s},\vec{y} \rangle \right]_{(K+1)\times 1}
\end{equation}
}\normalsize
The above shows the {existence} of the critical points.  It is easy to check that the critical point can only be the minimizer of 
the squared error $\mathcal{E}_{\mathbf{\Theta}}\left(\mathbf{x};g\right)$. Furthermore, we immediately have  
$\mathcal{E}(g_{\mathbf{\Theta}^{*}})\leq \min_{j\in[K]^{+}}\{\mathcal{E}(f_j)\}$.
\end{proof}

From the above proof, we can compute the minimizer for the case of the linear activation.
\begin{corollary}\label{FormulaSolution} 
The closed form of the minimizer is: 
\footnotesize{$$
\mathbf{\Theta}^{*}=\left[ \mathbf{\theta}_j\right]_{(K+1)\times 1} 
= 
\left[ \langle \vec{f_i},\vec{f_j} \rangle \right]_{(K+1)\times(K+1)}^{-1}\times
\left[ \langle \vec{f_j},\vec{y} \rangle \right]_{(K+1)\times 1}.
$$} 
\end{corollary}
{ }
{ }
Based on Lemma \ref{lemma_JL}, we can prove our next lemma

\begin{proof} (of Lemma \ref{lemma_3}) \\
Apply Lemma \ref{lemma_JL} to the given $\vec{y}$ and randomly selected $\vec{f}$, then we have
$$\Pr_{\vec{f}\in\mathbb{R}^{N}}\left\{ | \angle_{\vec{y},\vec{f}} -\frac{\pi}{2}|\leq \eta\right\} \geq 1- \frac{1}{\sqrt{N}}.$$ 
Also note that vectors $\vec{y}$, $\vec{f}$ and $\vec{f}-\vec{y}$ form a triangle with the three inner angles $\angle_{\vec{y} ,\vec{f}}$ , $\angle_{\vec{f}-\vec{y},\vec{f}}$ and $\angle_{\vec{f}-\vec{y},-\vec{y}}$, which means $\angle_{\vec{y},\vec{f}}+\angle_{\vec{f}-\vec{y},\vec{f}}+\angle_{\vec{f}-\vec{y},-\vec{y}}=\pi$. 
Hence, for large $N$,
$$\aligned
&\angle_{\vec{y},\vec{f}}=\frac{\pi}{2} \Rightarrow \angle_{\vec{f}-\vec{y},\vec{f}}\neq\frac{\pi}{2} \\
&\Rightarrow \Pr\left\{ \angle_{\vec{y},\vec{f}}=\frac{\pi}{2} \right\}\leq 
 \Pr\left\{ \angle_{\vec{f}-\vec{y},\vec{f}}\neq\frac{\pi}{2} \right\} \\
&\Rightarrow \Pr\left\{ \angle_{\vec{y},\vec{f}}\approx\frac{\pi}{2} \right\}\leq 
 \Pr\left\{ \angle_{\vec{f}-\vec{y},\vec{f}}\not\approx\frac{\pi}{2} \right\} \\
&\Rightarrow 1-\frac{1}{\sqrt{N}}\leq \Pr\left\{ \angle_{\vec{y},\vec{f}}\approx\frac{\pi}{2} \right\}\leq 
 \Pr\left\{ \angle_{\vec{f}-\vec{y},\vec{f}}\not\approx\frac{\pi}{2} \right\} \\
\endaligned $$ 
This means there exists a small enough $\eta>0$ s.t. 
$$\aligned
&1-\frac{1}{\sqrt{N}}\leq \Pr\left\{ |\angle_{\vec{y},\vec{f}}-\frac{\pi}{2}|\leq \eta \right\}\leq 
 \Pr\left\{ 
 |\angle_{\vec{f}-\vec{y},\vec{f}}-\frac{\pi}{2}|\geq \eta 
 \right\} \\ 
&\Rightarrow \frac{1}{\sqrt{N}} >  \Pr\left\{ |\angle_{\vec{f}-\vec{y},\vec{f}}-\frac{\pi}{2}|<\eta \right\}
\endaligned $$

In short, as $\angle_{\vec{f}-\vec{y},\vec{f}}$ is likely $\pi/2$, $\angle_{\vec{y},\vec{f}}$  must be less likely a vertical angle.   
Hence, $1-\frac{1}{\sqrt{N}}\leq\Pr\{|\angle_{\vec{f}-\vec{y},\vec{f}}-\frac{\pi}{2}|\leq\eta \} \leq \Pr\{| \angle_{\vec{y},\vec{f}} -\frac{\pi}{2}|>\eta \}.$
This completes the proof.
\end{proof}
{ }
{ }
\begin{proof} (of Lemma \ref{lemma_4}) \\
Observe that as $j^*$ is fixed and known,
$$\aligned
&\Pr\left\{\nabla_{ \mathbf{\Theta} } \mathcal{E}|_{\Theta^*=\vec{e_{j^*}}}
={\vec{0}} \right\}\\
&=\Pr\left\{\langle\vec{f}_{j^*}-\vec{y},\vec{f}_0 \rangle=0 \wedge\cdots\wedge
\langle\vec{f}_{j^*}-\vec{y},\vec{f}_K \rangle =0   \right\} \\
&\leq \Pr\left\{ \langle\vec{f}_{j^*}-\vec{y},\vec{f}_{j^*} \rangle =0 \right\}\\
&<\frac{1}{\sqrt{N}}
\endaligned $$
The last inequality is from Lemma \ref{lemma_3}.
However, in general $j^*$ is unknown, 
$$\aligned
&\Pr\left\{\exists\mathbf{\Theta}^*:\mathcal{E}(g_{\mathbf{\Theta}^{*}})= \min_{j\in[K]^{+}}\{\mathcal{E}(f_j)\} \right\} \\
&=\Pr\left\{\exists j \in [K]^+ s.t. \nabla_{ \mathbf{\Theta} } \mathcal{E}|_{\Theta^*=\vec{e_{j}}}
={\vec{0}} \right\}\\
&\leq \Pr\left\{ \vee_{j=0}^{K}\left\{ \langle\vec{f}_j-\vec{y},\vec{f}_j \rangle=0\right\} \right\}  \\
&= (K+1)\Pr\left\{ \langle\vec{f}-\vec{y},\vec{f} \rangle =0 \right\} \\
&<\frac{K+1}{\sqrt{N}}
\endaligned $$
Hence,
$$
\Pr\left\{\exists\mathbf{\Theta}^*\in\mathbb{R}^{K+1}s.t.\mathcal{E}(g_{\mathbf{\Theta}^{*}})< \min_{j\in[K]^{+}}\{\mathcal{E}(f_j)\} \right\}
>1-\frac{K+1}{\sqrt{N}}
$$
\end{proof}
{ }

\begin{proof} (of Lemma~\ref{lemma_case2}) \\
\textbf{For Eq. (8):} 
We first give a procedure of obtaining $g_{\mathbf{\Theta}_{\epsilon}}(\mathbf{x}^{(i)})$, then verify these settings in the procedure fit the conclusion of the first part: $\forall i\in[N]$, 
$|g_{\mathbf{\Theta}_{\epsilon}}(\mathbf{x}^{(i)})-g_{\mathbf{\Theta}^*_{0}}(\mathbf{x}^{(i)})|<\epsilon$. \\
\textbf{Procedure for Eq. (8):}\\
For the given $\epsilon$ and $\sigma(\cdot)$, we first find the following items based on the conclusions of Case 1 and Lemmas: \\ 
$g_{\mathbf{\Theta}^*_0}(\cdot)$. (By case 1)\\ 
$z_0\in \mathbb{R}$ s.t. $\frac{d}{dz}\sigma(z)\neq 0$. (By A3)\\
$U$ contains $z_0$. (By Lemma \ref{Inverse Function}) \\
$V$ contains $y_0$. (By Lemma \ref{Inverse Function}) \\
$\tau: V\to U$ s.t. $\forall z\in U$, $\tau(\sigma(z))=z$. (By Lemma \ref{Inverse Function}) \\
(Denote $y_0=\sigma(z_0)$, so $\tau(y_0)=z_0$.)\\
Then compute:
$$ \aligned
M_{g} & = \max\left\{1, \max_{i\in[N]}\{2\cdot|g_{\mathbf{\Theta}^*_0}(\mathbf{x}^{(i)})| \} \right\} \\
M_{\sigma} & = \max\left\{1,\sup_{z\in U}\{2\cdot\left(\frac{\sigma(z)-\sigma(z_0)}{z-z_0}\right)^2\} \right\}\\
M_{\tau} & = \max\left\{1,\sup_{z\in U}\{ |\tau^{(2)}(\sigma(z)-\sigma(z_0))| \} \right\} \\
M_{\gamma} & = \lceil\log_{2}(M_{g}M_{\sigma}M_{\tau}\epsilon^{-1}) \rceil+1 \\
\gamma_0 & = \sup_{z\in U}\left\{r=|z-z_0|: (z_0-r,z_0+r)\subset U \right\} \\
\gamma & = \min\left\{ \gamma_0, 2^{-M_{\gamma}} \right\}   \\
M_0 & =  \gamma^{-1} M_{g} \\
M_1 & = M_{\sigma}M_{\tau} 
\endaligned $$
Define: \\
$$ \aligned
L_{(0),\epsilon}(\mathbf{x}) &= M_0^{-1}g_{\mathbf{\Theta}^*_0}(\mathbf{x})+z_0 \\
L_{(1),\epsilon}(y) &= M_0\cdot\tau^{(1)}(y_0)\cdot y+
M_0\cdot\left(z_0-\tau^{(1)}(y_0)\cdot y_0 \right)\\
\endaligned $$
\textbf{Verification}: \\ 
First observe that $L_{(0),\epsilon}(\mathbf{x})$ is a linear combination with a bias, i.e., an affine mapping, since $g_{\mathbf{\Theta}^*_0}(\mathbf{x})$ itself is an affine mapping. 
Similarly,  $L_{(1),\epsilon}(y)$ is an affine mapping of $y$.

Next, for all $i\in[N]$, $L_{\Theta_{0,\epsilon}}(\mathbf{x}^{(i)})=M_0^{-1}g_{\mathbf{\Theta}^*_0}(\mathbf{x}^{(i)})+z_0\in (-\gamma+z_0,z_0+\gamma)\subset U$ since $\gamma\leq \frac{\gamma_0}{2}$ and $(-\frac{\gamma_0}{2}+z_0,z_0+\frac{\gamma_0}{2})\subset U $.
Hence, by Lemma \ref{Inverse Function}, $$
\tau\left(\sigma\left(L_{\Theta_{0,\epsilon}}(\mathbf{x}^{(i)}) \right) \right)= L_{\Theta_{0,\epsilon}}(\mathbf{x}^{(i)}).
$$ 

Now let $z\in (-\gamma+z_0,z_0+\gamma)$ and $y=\sigma(z)$, then by Lemma \ref{TaylorLagrange} and Eq. (\ref{eq_8}), 
$$\aligned 
& |\tau\left(y\right)-\left(\tau(y_0)+\tau^{(1)}(y_0)(y-y_0)\right)|\\
 & = \frac{\tau^{(2)}(c(y-y_0))}{2!} (y-y_0)^2 \\
 &<  2\cdot\sup_{z\in U}\left\{ |\tau^{(2)}(\sigma(z)-\sigma(z_0))| \cdot\left(\frac{\sigma(z)-\sigma(z_0)}{z-z_0}\right)^2\right\}\cdot(z-z_0)^2 \\
 & \leq M_{\tau}M_{\sigma}\gamma^2 =M_1\gamma^2
\endaligned $$ 
Replace $y$ with $\sigma(z)$ and simplify the expression in the absolute value symbol, then we have $\tau\left(y\right)=\tau\left(\sigma(z)\right)=z$. Furthermore,  
$\tau(y_0)+\tau^{(1)}(y_0)(y-y_0) = \tau^{(1)}(y_0)\cdot y+ \left(\tau(y_0)-\tau^{(1)}(y_0)\cdot y_0 \right)$. 
Then replace $z$ with $L_{\Theta_{0,\epsilon}}(\mathbf{x}^{(i)})$, and $\tau(y_0)$ with $z_0$,   
$$\aligned 
& | M_0^{-1}g_{\mathbf{\Theta}^*_0}(\mathbf{x})+z_0 -
\left\{ z_0 +\tau^{(1)}(y_0)\left(\sigma\left( L_{\Theta_{0,\epsilon}}(\mathbf{x}^{(i)}) \right)-y_0 \right) \right\} |\\
& < M_1\gamma^2
\endaligned $$ \\
This means that
$$\aligned 
&|g_{\mathbf{\Theta}^*_0}(\mathbf{x})-L_{\Theta_{1,\epsilon}}\left(\sigma\left( L_{\Theta_{0,\epsilon}}(\mathbf{x}^{(i)}) \right)\right)|<  M_0M_1\gamma^2 \\
& \Rightarrow
|g_{\mathbf{\Theta}^*_0}(\mathbf{x})-g_{\mathbf{\Theta}_{\epsilon}}(\mathbf{x})|<  M_0M_1\gamma^2 \\
\endaligned $$ 

Recall that $\gamma\leq 2^{-M_{\gamma}}<\frac{\epsilon}{M_{g}M_{\sigma}M_{\tau}}$. Hence,
$$
M_0M_1\gamma^2=\gamma^{-1}M_{g}M_{\sigma}M_{\tau}\gamma^2
=M_{g}M_{\sigma}M_{\tau}\gamma
<\epsilon
$$
achieve the goal of the first part of this Lemma.

\textbf{For Eq. (9):} 
For the second part, we claim the following settings satisfy $\mathcal{E}(g_{\mathbf{\Theta}_{\epsilon}})\leq \frac{2\mathcal{E}(g_{\mathbf{\Theta}^{*}_0})+\mathcal{E}(f_{j^*})}{3}
<\mathcal{E}(f_{j^*})$. \\
\textbf{Procedure for Eq. (9):\\}
Compute and then set these:
$$\aligned
&M_2=\max_{i\in[N]}\left\{|g_{\mathbf{\Theta}^{*}_0}(\mathbf{x}^{(i)})-y^{(i)}|\right\}\\
&\epsilon=\frac{\mathcal{E}(f_{j^*})-\mathcal{E}(g_{\mathbf{\Theta}^{*}_0})}{4N(2M_2+1)}
\endaligned$$
\textbf{Verification:} \\
Observe that 
$$\aligned
&\mathcal{E}(g_{\mathbf{\Theta}^{*}_0})< \mathcal{E}(f_{j^*})
\Rightarrow \\
&\max_{i\in[N]}\left\{(f_{j^*}(\mathbf{x}^{(i)})-y^{(i)})^2-(g_{\mathbf{\Theta}^{*}_0}(\mathbf{x}^{(i)})-y^{(i)})^2\right\}>0 \\
&\mathcal{E}(g_{\mathbf{\Theta}^{*}_0})+ \frac{\mathcal{E}(f_{j^*})-\mathcal{E}(g_{\mathbf{\Theta}^{*}_0})}{3} 
= \frac{2\mathcal{E}(g_{\mathbf{\Theta}^{*}_0})+\mathcal{E}(f_{j^*})}{3}
<\mathcal{E}(f_{j^*}) 
\endaligned$$
Besides,
$$N\cdot(2M_2+1)\cdot\epsilon<\frac{\mathcal{E}(f_{j^*})-\mathcal{E}(g_{\mathbf{\Theta}^{*}_0})}{3} $$
and
$$\aligned
&|g_{\mathbf{\Theta}_{\epsilon}}(\mathbf{x})-g_{\mathbf{\Theta}^*_{0}}(\mathbf{x})|<\epsilon \\
&\Rightarrow |(g_{\mathbf{\Theta}_{\epsilon}}(\mathbf{x})-y)-(g_{\mathbf{\Theta}^*_{0}}(\mathbf{x})-y)|<\epsilon\\
&\Rightarrow 0\leq |g_{\mathbf{\Theta}_{\epsilon}}(\mathbf{x})-y|<|g_{\mathbf{\Theta}^*_{0}}(\mathbf{x})-y|+\epsilon \\
&\Rightarrow (g_{\mathbf{\Theta}_{\epsilon}}(\mathbf{x})-y)^2 < (|g_{\mathbf{\Theta}^*_{0}}(\mathbf{x})-y|+\epsilon )^2
\endaligned$$
Hence, based on above observations we have 
$$\aligned
\mathcal{E}(g_{\mathbf{\Theta}_{\epsilon}})
=&\sum_{i\in[N]}{(g_{\mathbf{\Theta}_{\epsilon}}(\mathbf{x}^{(i)})-y^{(i)})^2} \\
<&\sum_{i\in[N]}\{|g_{\mathbf{\Theta}_0^{*}}(\mathbf{x}^{(i)})-y^{(i)}|+\epsilon\}^2 \\
=&\sum_{i\in[N]}(g_{\mathbf{\Theta}_0^{*}}(\mathbf{x}^{(i)})-y^{(i)})^2 \\& +\sum_{i\in[N]}\left\{2\epsilon\cdot|g_{\mathbf{\Theta}_0^{*}}(\mathbf{x}^{(i)})-y^{(i)}|+\epsilon^2  \right\}\\
= &\mathcal{E}(g_{\mathbf{\Theta}_0^{*}})+\epsilon\cdot\sum_{i\in[N]}\left(2|g_{\mathbf{\Theta}_0^{*}}(\mathbf{x}^{(i)})-y^{(i)}|+\epsilon\right)\\
\leq & \mathcal{E}(g_{\mathbf{\Theta}_0^{*}})+\epsilon\cdot N\cdot\left(2M_2+1\right) \\
< & \mathcal{E}(g_{\mathbf{\Theta}^{*}_0})+ \frac{\mathcal{E}(f_{j^*})-\mathcal{E}(g_{\mathbf{\Theta}^{*}_0})}{3} \\
= & \frac{\mathcal{E}(f_{j^*})+2\mathcal{E}(g_{\mathbf{\Theta}^{*}_0})}{3} \\
< & \mathcal{E}(f_{j^*})
\endaligned $$
which means that $\mathcal{E}(g_{\mathbf{\Theta}_{\epsilon}})< \min_{j\in[K]^{+}}\{\mathcal{E}(f_j)\}$. The proof is complete.
\end{proof}

\begin{proof}(of Proposition \ref{AddOne_Kis2})\\
Let 
$$\aligned
 &D(\alpha_0,\alpha_1)\\
 &=\sum_{i\in[N]}{(f_1(\mathbf{x}^{(i)})-y^{(i)})^2-\left(\alpha_0 f_0(\mathbf{x}^{(i)})+\alpha_1 f_1(\mathbf{x}^{(i)})-y^{(i)}\right)^2}.
\endaligned $$
First observe that $D(0,1)=0$ and hence if $\nabla D(0,1)\neq (0,0)$ then it is easy to know that $\exists (\alpha_0^{*},\alpha_1^{*})$ s.t. $D(\alpha_0^{*},\alpha_1^{*})>0$.
$$
\nabla D(\alpha_0,\alpha_1)=
-2\cdot 
\begin{bmatrix}
\langle \alpha_0 \vec{f}_0 +\alpha_1 \vec{f}_1-\vec{y},\vec{f}_0 \rangle \\
\langle \alpha_0 \vec{f}_0 +\alpha_1 \vec{f}_1-\vec{y},\vec{f}_1 \rangle
\end{bmatrix} 
$$
Then, by considering $(\alpha_0,\alpha_1)=(0,1)$ we have
$$
\nabla D(0,1)=
-2\cdot 
\begin{bmatrix}
\langle  \vec{f}_1-\vec{y},\vec{f}_0 \rangle \\
\langle  \vec{f}_1-\vec{y},\vec{f}_1 \rangle
\end{bmatrix} 
$$
Apply Lemma \ref{lemma_3},  
$$\aligned
&\Pr\left\{\nabla D|_{\Theta^*=\vec{e_{j^*}}}
={\vec{0}} \right\}\\
&\leq \Pr\{\exists j\in[1]^{+} s.t.\langle \vec{f}_j-\vec{y},\vec{f}_j\rangle=0 \}\\ &<\frac{2}{\sqrt{N}}
\endaligned$$
That is, 
$$\aligned
& \Pr\{\exists (\alpha_0,\alpha_1) s.t. D(\alpha_0,\alpha_1)>0 \} \\
& \geq \Pr\{\nabla D(0,1)\neq \vec{0} \} \\
& >1-\frac{2}{\sqrt{N}}
\endaligned$$
\end{proof}

\begin{proof}(of Lemma \ref{AddOne})\\
We first prove this lemma of linear activation, and then similar to previous section apply Lemma \ref{lemma_case2} to address the non-linear activation. For the linear activation, it can be proved by induction.

\textbf{Base case:} It is done in Proposition \ref{AddOne_Kis2}. 

\textbf{Inductive step:} Suppose as $J=k-1$ the statement is true. That is, $g_{k-1}=L_{\Theta}(f_1,...,f_{k-1})$ and with probability at least $1-\frac{K}{\sqrt{N}}$
, there is $\mathbf{\Theta}$ s.t.  $\mathcal{E}\left(g_{K-2}\right)> \mathcal{E}_{\mathbf{\Theta}}\left(g_{K-1}\right)$.
As $J=k$, let $f_0$ and $f_1$ in Proposition \ref{AddOne_Kis2} be $g_{k-1}$ and $f_k$ respectively. Then we have  $\alpha_0g_{k-1}+\alpha_1 f_k$ as the composite network. 
Repeat the argument in the previous proposition, then we can conclude with probability at least $1-\frac{k+1}{\sqrt{N}}$ there is $(\alpha_0,\alpha_1)$ s.t.  $\mathcal{E}\left(g_{K-1}\right)> \mathcal{E}_{\mathbf{\Theta}}\left(\alpha_0g_{k-1}+\alpha_1 f_k\right)$. Note that $\alpha_0g_{k-1}+\alpha_1 f_k$ is a possible form of $g_{K}$. So the statement holds. The details are as follows:  
\scriptsize{$$\aligned
 &D(\alpha_0,\alpha_1)\\
 &=\sum_{i\in[N]}{(g_{k-1}(\mathbf{x}^{(i)})-y^{(i)})^2-\left(\alpha_0 g_{k-1}(\mathbf{x}^{(i)})+\alpha_1 f_{k}(\mathbf{x}^{(i)})-y^{(i)}\right)^2}.
\endaligned $$}\normalsize 
First observe that $D(1,0)=0$ and hence if $\nabla D(1,0)\neq \vec{0} $ then it is easy to know that $\exists (\alpha_0^{*},\alpha_1^{*})$ s.t. $D(\alpha_0^{*},\alpha_1^{*})>0$.
$$
\nabla D(\alpha_0,\alpha_1)=
-2\cdot 
\begin{bmatrix}
\langle \alpha_0 \vec{g}_{k-1} +\alpha_1 \vec{f}_k-\vec{y},\vec{g}_{k-1}\rangle \\
\langle \alpha_0 \vec{g}_{k-1} +\alpha_1 \vec{f}_k-\vec{y},\vec{f}_k \rangle 
\end{bmatrix} 
$$
Then, 
$$
\nabla D(1,0) =
-2\cdot 
\begin{bmatrix}
\langle  \vec{g}_{k-1}-\vec{y},\vec{g}_{k-1} \rangle \\
\langle  \vec{g}_{k-1}-\vec{y},\vec{f}_k \rangle)
\end{bmatrix} 
$$
Apply Lemma \ref{lemma_4} and by Induction hypothesis, we have 
$$\aligned
& \Pr\left\{\nabla D|_{\Theta^*=\vec{e_{j^*}}}
={\vec{0}} \right\}\\
&\leq \Pr\{\langle \vec{g}_{k-1}-\vec{y},\vec{g}_{k-1} \rangle=0\} +\Pr\{\langle \vec{f}_{k}-\vec{y},\vec{f}_{k} \rangle=0 \}\\ 
&<\frac{k}{\sqrt{N}}+\frac{1}{\sqrt{N}}=\frac{k+1}{\sqrt{N}}
\endaligned$$
Thus, 
$$\aligned
& \Pr\{\exists (\alpha_0,\alpha_1) s.t. D(\alpha_0,\alpha_1)>0 \} \\
& \geq \Pr\left\{\nabla D|_{\Theta^*=\vec{e_{j^*}}}
\neq {\vec{0}} \right\} \\
& >1-\frac{k+1}{\sqrt{N}}
\endaligned$$
This completes the inductive step.

For the non-linear activation, repeat the argument of Lemma 7 to obtain a proper $g_{\mathbf{\Theta}_{\epsilon}}$ corresponding to the given $\epsilon$ and the linear mapping $g_{\mathbf{\Theta}_{0}^{*}}$, and a small enough $\epsilon$ can yield a proper $\mathbf{\Theta}_{\epsilon}$ that fits the conclusion of $\mathcal{E}(g_{K-1})>\mathcal{E}_{\mathbf{\Theta}_{\epsilon}}(g_{K})$.
The probability of existence is inherently obtained as the same as in Lemma 7. 
\end{proof}

\begin{proof}(of Lemma \ref{addDeep})\\
Observe that for the given set of pre-tained components $\{f_j\}_{j\in[K]}$ and by the definition of $g_{K-1}$, $f_K$ is not a component of $g_{K-1}$. Hence, if the activation functions used in the construction of $g_{K-1}$ are all linear, the assumption A1 implies that $\vec{g}_{K-1}$ is linear independent of  $\vec{f}_{K}$. Furthermore, if there is at least one non-linear activation function used in the construction of $g_{K-1}$, then as $N$ is large enough, Lemma \ref{lemma_JL} implies that $\vec{g}_{K-1}$ and $\vec{f}_{K}$ are not parallel with a very high probability. This means the assumption that $\vec{g}_{K-1}$ is linear independent of  $\vec{f}_{K}$ is reasonable. Furthermore, this implies that the events $E_1:\exists\mathbf{\Theta}s.t.\mathcal{E}_{\mathbf{\Theta}}(g_{K})<min\{\mathcal{E}(g_{k-1}),\mathcal{E}(f_{k})\}$, and $E_2:\mathcal{E}(g_{K-1})<\cdots< \min_{j\in[K]^{+}}\{\mathcal{E}(f_j)\}$, are independent. Hence, $\Pr\{E_1|E_2\}=\Pr\{E_1\}$.
\tiny
\normalsize
\end{proof}

\begin{proof}(of Theorem \ref{DeeperWider})\\
For a set of given $K$ pre-trained components, 
$g_{k}:=L_{(k)}(\sigma(L_{(k-1)}(\cdots L_{(1)}(\sigma(L_{(0)}(f_1,\cdots,f_K)))\cdots)))$ is one of possible 
$H$-hidden layer composite network. Hence obviously, 
$$\aligned
&\Pr\left\{\exists\mathbf{\Theta}^{*}:\mathcal{E}(g_{\mathbf{\Theta}^{*}})< \min_{j\in[K]^{+}}\{\mathcal{E}(f_j)\} \right\}\\
&\geq \Pr\left\{\mathcal{E}(g_{H})<\mathcal{E}(g_{H-1})<\cdots<\mathcal{E}(g_{1})< \min_{j\in[K]^{+}}\{\mathcal{E}(f_j)\} \right\}\\
&\geq \Pr\left\{\mathcal{E}(g_{1})< \min_{j\in[K]^{+}}\{\mathcal{E}(f_j)\} \right\} \\ &\mbox{ }\times \Pr\left\{\mathcal{E}(g_{2})<\mathcal{E}(g_{1})\mid\mathcal{E}(g_{1})< \min_{j\in[K]^{+}}\{\mathcal{E}(f_j)\} \right\}\times \cdots \times \\ 
&\mbox{ } \Pr\left\{\mathcal{E}(g_{H})<\mathcal{E}(g_{H-1})\mid \mathcal{E}(g_{H-1})<\cdots< \min_{j\in[K]^{+}}\{\mathcal{E}(f_j)\} \right\}\\
&= \left(1-\frac{K+1}{\sqrt{N}}\right)^H
\endaligned 
$$
The last inequality is based on Lemmas 8 and 9: 
$$
\min_{k\in[H]}\left\{ P_k \right\} \geq 1-\frac{K+1}{\sqrt{N}},
$$
where \small{
$$\aligned
P_k & =\Pr\left\{\exists\mathbf{\Theta}:\mathcal{E}(g_{k})<\mathcal{E}(g_{k-1})\mid \mathcal{E}(g_{k-1})<\cdots< \min_{j\in[K]^{+}}\{\mathcal{E}(f_j)\} \right\} \\
  & =\Pr\left\{\exists\mathbf{\Theta}:\mathcal{E}(g_{k})<\mathcal{E}(g_{k-1})\right\} \mbox{ by Lemma 9.}
\endaligned$$
}\normalsize
This completes the proof.
\end{proof}

\newpage
\textbf{A2. More Details of Experiments}\\
\scriptsize{
In Table \ref{tab: Architecture and Hyperparameters of Components} shows the details of each component:
\begin{table}[h]
    \centering
    \caption{Architectures and Hyperparameters of Components}\label{tab: Architecture and Hyperparameters of Components}
    \begin{tabular}{r|l}
    \toprule
    Comp. & \textbf{Descriptions}: attention (decoder) layer (Att),\\
     & time length $l\in \{24,48,72\}$,  batch normalization (BN),\\
          & convolutional layer (Cvl), max-pooling (MaxP),   \\
          & flatten layer (Fltn), dense layer (Den), dropout (Drop)\\
    \midrule
     $f_1$    & Input layer: $(30\times 38) \times 9 $ \\
         &  hidden layers: BN, Cvl-1($30\times 38$, 32 filters), MaxP, \\
         & \mbox{ }\mbox{ }\mbox{ } Cvl-2($15\times 19$, 16 filters), MaxP, Fltn, \\
         & \mbox{ }\mbox{ }\mbox{ } LSTM(150, time $l$), Att, Fltn, Den, Drop(0.2), Den \\
         & Output layer: Den(RELU), 18\\
         & total parameters: 917,510\\
    \midrule     
      $f_2$  & Input layer: $(30\times 38) \times 4 $   \\
          &  hidden layers:  same with $f_1$\\
         & Output layer: Den(RELU), 18\\
         & total parameters: 916,918\\
    \midrule     
      $f_3$  & Input layer: $(30\times 38) \times 9 $   \\
          &  hidden layers: Fltn, BN, Den-1(100), Den-2(100)\\
         & Output layer: Den(RELU), 18\\
         & total parameters: 1,038,054\\
    \midrule     
      $f_4$  &  Input layer: $(30\times 38) \times 4 $   \\
          &  hidden layers: same with $f_3$ \\
         & Output layer: Den(RELU), 18\\
         & total parameters: 582,038\\
    \midrule     
     $f_5$   & Input layer: one-hot $(24+7+12)$ \\
         &  hidden layers: LSTM(150, time $l$), Att, Fltn, Den, \\
         &  \mbox{ }\mbox{ }\mbox{ } Drop(0.2), Den \\
         & Output layer: Den(Scaled-Logistic), 18\\
         & total parameters: 953,916\\
    \midrule     
       $f_{W_6}$  &  Input layer: $(30\times 38) \times 4$  \\
         &  hidden layers: Cvl-1($30\times 38$, 16 filters)\\
         & \mbox{ }\mbox{ }\mbox{ } Cvl-2($15\times 19$, 16 filters), MaxP, Fltn, \\
         & Output layer: Den(RELU), 18\\
         & total parameters: 41,380\\
    \bottomrule     
    \end{tabular}
\end{table}
}\normalsize

\begin{table}[ht]
\centering \scriptsize{
\caption{Composite networks by Algo 1: DBCN, Next 48hr.} \label{Table_Algo1_48hr}
\begin{tabular}{l|rrr|l}  
\toprule
      &\multicolumn{2}{c}{RMSE}  & parameter & \\
     \cmidrule(r){2-3}
Model &Training & \textbf{Testing} &trainable/total & note\\
\midrule
$g_1\leftarrow f_1$& & & 0/ &  \\
$L(g_1,f_5)$  & 8.1850 & \textbf{11.0995}  &666/1872092 & $g_2$\\  
$SL(g_1,f_5)$  & 8.1675 & 11.2399&666/1872092& \\ 
\midrule
$L(g_2,f_4)$  & 8.1520 & 11.2632   &666/2454796 & \\  
$SL(g_2,f_4)$  & 8.1902 & \textbf{11.1647} &666/2454796& $g_3$\\ 
\midrule
$L(g_3,f_3)$  & 8.1382 & \textbf{11.1163}  &666/3493516 & $g_4$\\  
$SL(g_3,f_3)$  & 8.1085 & 11.1442   &666/3493516 & \\ 
\midrule
$L(g_4,f_2)$  & 8.0361 & 11.0991 &666/4411100 & \\  
$SL(g_4,f_2)$  & 8.0678 & \textbf{11.0469} &666/4411100& $g_5$\\ 
\midrule
$L(g_5,f_{W_6})$  & 7.8941 & \textbf{10.9531}  &42046/4453146 & $g_6$\\  
$SL(g_5,f_{W_6})$  & 7.9009 & 10.9754 &42046/4453146& \\ 
\bottomrule
\end{tabular} }
\end{table} \normalsize 

\begin{table}[ht]
\centering \scriptsize{ 
\caption{Composite networks by Algo 1: DBCN, Next 72hr.} \label{Table_Algo1_72hr}
\begin{tabular}{l|rrr|l}  
\toprule
      &\multicolumn{2}{c}{RMSE}  & parameter & \\
     \cmidrule(r){2-3}
Model &Training & \textbf{Testing} &trainable/total & note\\
\midrule
$g_1\leftarrow f_5$ & & & &  \\
$L(g_1,f_1)$  & 8.4308 & 11.3572  &666/1872092 & \\  
$SL(g_1,f_1)$  & 8.2979 & \textbf{11.3323}  &666/1872092& $g_2$\\ 
\midrule
$L(g_2,f_2)$  & 8.3579 & \textbf{11.3634}  &666/2789676 &  $g_3$\\  
$SL(g_2,f_2)$  & 8.3252 & 11.4001  &666/2789676&\\ 
\midrule
$L(g_3,f_4)$  & 8.4116 & \textbf{11.4195} &666/3372380 & $g_4$\\  
$SL(g_3,f_4)$  & 8.6230 & 11.4530  &666/3372380& \\ 
\midrule
$L(g_4,f_3)$  & 8.2305 & \textbf{11.4274}   &666/4411100& $g_5$\\  
$SL(g_4,f_3)$  & 8.1284 & 11.4482 &666/4411100& \\ 
\midrule
$L(g_5,f_{W_6})$  & 8.2448 & \textbf{11.2541} & 42046/4453146 & $g_6$\\  
$SL(g_5,f_{W_6})$  & 8.2125 & 11.3232  &42046/4453146 & \\ 
\bottomrule
\end{tabular} }
\end{table} \normalsize 

\newpage
\begin{table}[ht]
\centering \scriptsize{
\caption{Composite networks by Algo 2: BBCN, Next 48hr.} \label{Table_Algo2_48hr}
\begin{tabular}{l|rrr|l}  
\toprule
      &\multicolumn{2}{c}{RMSE}& parameter & \\
      \cmidrule(r){2-3}
Model &Training & \textbf{Testing} &trainable/total & note\\
\midrule
\multicolumn{5}{l}{$h_{0,1}\leftarrow f_1,h_{0,2}\leftarrow f_2$}  \\
$L(h_{0,1},h_{0,2})$  & 6.1001 & 11.1004 & 666/1835094& \\  
$SL(h_{0,1},h_{0,2})$  &5.5894 & \textbf{11.0907}   &666/1835094 & $h_{1,1}$\\ 
\midrule
\multicolumn{5}{l}{$h_{0,3}\leftarrow f_3,h_{0,4}\leftarrow f_4$ } \\
$L(h_{0,3},h_{0,4})$  & 11.8311 & \textbf{11.5098}& 666/1620758 & $h_{1,2}$\\  
$SL(h_{0,3},h_{0,4})$  & 11.6587 & 11.5436 &666/1620758 &\\ 
\midrule
$L(h_{1,1},h_{1,2})$  & 8.2277 & 11.1739  & 666/3456518 & \\  
$SL(h_{1,1},h_{1,2})$  & 7.9990 & \textbf{11.0935}  &666/3456518 & $h_{2,1}$\\ 
\midrule
\multicolumn{5}{l}{$h_{2,2}\leftarrow h_{1,3}\leftarrow h_{0,5}\leftarrow f_5$ } \\
$L(h_{2,1},h_{2,2})$  & 8.0273 & 11.0808  & 666/4411100 & \\  
$SL(h_{2,1},h_{2,2})$  & 7.9949 & \textbf{11.0516} &666/4411100  & $h_{3,1}$ \\ 
\midrule
\multicolumn{5}{l}{$g_5\leftarrow h_{3,1}$} \\
$L(g_5,f_{W_6})$  & 8.5736 & \textbf{11.0182}  & 42046/4453146& $g_6$\\  
$SL(g_5,f_{W_6})$  & 7.7346 & 11.0208  & 42046/4453146 &\\ 
\bottomrule
\end{tabular} }
\end{table} \normalsize 

\begin{table}[ht]
\centering \scriptsize{ 
\caption{Composite networks by Algo 2: BBCN, Next 72hr.} \label{Table_Algo2_72hr}
\begin{tabular}{l|rrr|l}  
\toprule
      &\multicolumn{2}{c}{RMSE}& parameter & \\
      \cmidrule(r){2-3}
Model &Training & \textbf{Testing} &trainable/total & note\\
\midrule
\multicolumn{5}{l}{$h_{0,1}\leftarrow f_1,h_{0,2}\leftarrow f_2$}\\
$L(h_{0,1},h_{0,2})$  &  7.7873 & 11.4480 & 666/1835094 & \\  
$SL(h_{0,1},h_{0,2})$  &7.9830 & \textbf{11.4198}   &666/1835094&$h_{1,1}$\\ 
\midrule
\multicolumn{5}{l}{$h_{0,3}\leftarrow f_3,h_{0,4}\leftarrow f_4$ }\\
$L(h_{0,3},h_{0,4})$  &  12.0284 & \textbf{11.7405}  &666/1620758& $h_{1,2}$\\  
$SL(h_{0,3},h_{0,4})$  & 12.0460 & 11.8005&666/1620758 &\\ 
\midrule
$L(h_{1,1},h_{1,2})$  & 8.3884 & 11.7030 & 666/3456518 & \\  
$SL(h_{1,1},h_{1,2})$  & 8.5149 & \textbf{11.5703}  &666/3456518 &$h_{2,1}$ \\ 
\midrule
\multicolumn{5}{l}{$h_{2,2}\leftarrow h_{1,3}\leftarrow h_{0,5}\leftarrow f_5$ } \\
$L(h_{2,1},h_{2,2})$  & 8.4460 & \textbf{11.5100}  & 666/4411100 & $h_{3,1}$\\  
$SL(h_{2,1},h_{2,2})$  & 8.4093 & 11.5526  &666/4411100&\\ 
\midrule
\multicolumn{5}{l}{$g_5\leftarrow h_{3,1}$}\\
$L(g_5,f_{W_6})$  & 9.1848 & \textbf{11.4153}  & 42046/4453146 & $g_6$\\  
$SL(g_5,f_{W_6})$  & 9.0706 & 11.4474  & 42046/4453146& \\ 
\bottomrule
\end{tabular} }
\end{table} \normalsize 


\begin{table}[ht]
\hfill{} 
 \centering
 \scriptsize{ 

\caption{Summary of all methods (MAE) }
  \label{Summary_all_methods_MAE_SMAPE}
\begin{tabular}{l|rrrrrr}  
\toprule
{}  & \multicolumn{2}{l}{+24h}   & \multicolumn{2}{l}{+48h}  & \multicolumn{2}{l}{+72h} \\
               \cmidrule(r){2-3} \cmidrule(r){4-5}\cmidrule(r){6-7}  
Method &{Training} &{Testing}  & {Training} & {Testing} &  {Training} & {Testing}   \\
\midrule
SVM & 8.0701 & \textbf{7.7026}  & 8.5887 & \textbf{8.3525} & 8.6831 & \textbf{8.6157}    \\ 
Random forests & 2.3956 & 8.3523  & 2.5007 & 9.2156 &  2.5131 &  9.4219 \\ 
Ensemble & 8.8470 & 8.5245  & 9.1113 & 8.7430 &  9.3899 &  8.8767    \\ 
\emph{SL}(Ensemble) & 8.7689 & 8.4413  & 9.0365 & 8.6116 &  9.3121 & 8.8657    \\ 
DBCN$_{ \mathsf{Relu}}$ & 9.3082 &8.7309 &9.8383 & 9.1429& 10.4367&9.6356 \\
DBCN$_{ \mathsf{Sigm}}$ &8.6927 & 8.2972& 10.0452&9.6134 &9.9240 &9.5408\\
DBCN & 3.7188  & \textbf{7.7868}  & 4.3161 & \textbf{8.4258} & 4.3261 & \textbf{8.8250} \\
BBCN$_{ \mathsf{Relu}}$ & 9.7099 &9.2274 &10.6558 &10.0856 & 11.6371& 10.5833\\
BBCN$_{ \mathsf{Sigm}}$ & 9.1872& 8.7346 & 9.7944& 9.5822& 9.6479& 8.8532\\
BBCN & 3.7676 & 7.9866 & 4.4403 & 8.5564 & 4.5324 & 8.8781 \\
Exhaustive-a   & 2.8646 & \textbf{6.8319} & 2.8078 & \textbf{7.6136} & 3.6612 &\textbf{7.7701}  \\
\midrule
\midrule
\multicolumn{7}{l}{(Include  $f_{W_6}$)} \\ 
Ensemble   &  8.7257 & 8.2739 & 8.9901 & \textbf{8.2476} & 9.2801 & 8.7263\\
\emph{SL}(Ensemble) &  8.7470 & 8.3988 & 9.0098 & 8.3065 & 9.2281 &  \textbf{8.4536} \\
DBCN$_{ \mathsf{Relu}}$ &9.3969 &8.5961 & 10.2851&9.5191 &10.4696 &9.4412 \\
DBCN$_{ \mathsf{Sigm}}$ &8.7275 &8.1512 &8.9281 &8.2625 & 9.6004&8.8777 \\
DBCN & 3.6608 & \textbf{7.5614}  & 4.2419 & 8.2766 & 4.4143 & 8.5776  \\
BBCN$_{ \mathsf{Relu}}$ &8.5198 &8.0122 &8.9259 &8.4752 &9.2290 &8.5274 \\
BBCN$_{ \mathsf{Sigm}}$ & 9.1376&8.6430 &9.6253 &9.0825 &9.6115 &8.8030 \\
BBCN & 3.6698 & 7.6156 & 4.6652 & 8.4528 & 4.8884 & 8.6097  \\
Exhaustive-a  & 3.1250 & \textbf{6.7757} & 2.8133 & \textbf{7.5986} & 4.5569 & \textbf{7.6798} \\
Exhaustive-b  & 3.8088 & \textbf{6.9032} & 3.1118 & \textbf{7.5156} & 3.0740 & \textbf{7.6561}\\
\bottomrule
\end{tabular} } 

\hfill{}
\end{table} 

\newpage
\begin{figure}[h]
    \centering
    \includegraphics[width=0.5\textwidth]{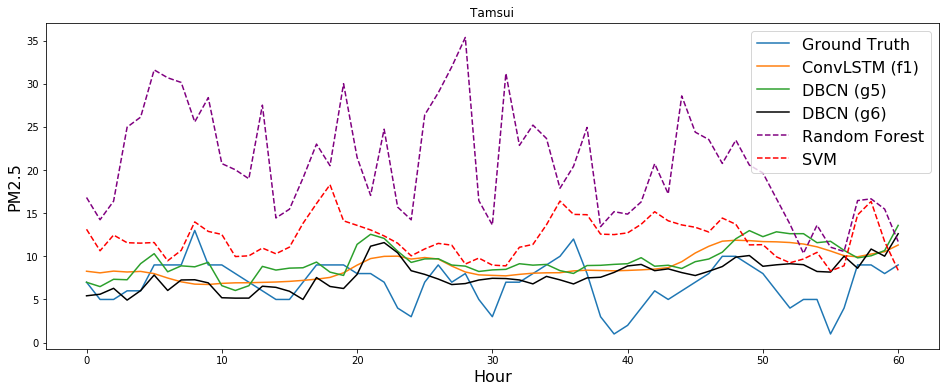}
    \caption{Case study for Tamsui station}
    \label{fig:Case_study_for_Tamsui_station}
\end{figure}

Figure \ref{fig:Case_study_for_Tamsui_station} illustrates
a typical example of the next-24-hour predictions of various models, including $g_5$ and $g_6$ of DBCN, ConvLSTM, SVM and random forest, of Tamsui for 60 hours starting from 9:00 pm, October 22, 2016. 
$f_1$ is a ConvLSTM model that its prediction 
is central to the average of the ground truth. 
DBCN ($g_6$) considers the one extra weather feature, i.e., the chance of rain in the future, that it produces a lower PM2.5 prediction than DBCN($g_5$).  In this duration, the traditional machine learning methods SVM and RF usually overestimated, although they apply the same weather and pollutant features.

\end{document}